\documentclass[11pt]{article}

\usepackage[final]{acl}

\usepackage{times}
\usepackage{latexsym}
\usepackage{multirow}
\usepackage[T1]{fontenc}
\usepackage[utf8]{inputenc}
\usepackage{microtype}
\usepackage{inconsolata}
\usepackage{graphicx}
\usepackage{amsmath}
\usepackage{amssymb}
\usepackage{algorithm}
\usepackage{algpseudocode}
\usepackage{booktabs}
\usepackage{xcolor}

\newcommand{\best}[1]{\textcolor{blue!70!black}{\textbf{#1}}}
\newcommand{\second}[1]{\underline{#1}}
\newcommand{\rev}[1]{#1}

\newcommand{\floatvspace}[1]{\vspace{#1}}
\usepackage{titlesec}
\titlespacing*{\section}{0pt}{6pt plus 2pt minus 2pt}{3pt plus 1pt minus 1pt}
\titlespacing*{\subsection}{0pt}{5pt plus 2pt minus 2pt}{2pt plus 1pt minus 1pt}
\titlespacing*{\paragraph}{0pt}{4pt plus 1pt minus 1pt}{0.5em}
\usepackage{enumitem}
\setlist{topsep=2pt,itemsep=1pt,parsep=0pt,partopsep=0pt}

\title{Self-Supervised On-Policy Distillation for Omni-Modal Video Captioning}

\author{
  \textbf{Zinuo Li}$^{1,3}$, \textbf{Yongxin Guo}$^{2}$, \textbf{Xin Li}$^{3}$,
  \textbf{Hao Tan}$^{3}$, \textbf{Zhao Zhong}$^{3}$, \\
  \textbf{Liefeng Bo}$^{3}$, \textbf{Mohammed Bennamoun}$^{1}$,
  \textbf{Farid Boussaid}$^{1}$, \textbf{Qiuhong Ke}$^{4}$ \\[2pt]
  $^{1}$ University of Western Australia \quad
  $^{2}$ Chinese University of Hong Kong (SZ) \\
  $^{3}$ Tencent Hunyuan \quad
  $^{4}$ Monash University \\
  {\definecolor{projectpagepink}{HTML}{C2185B}\hypersetup{urlcolor=projectpagepink}\urlstyle{same}\url{https://github.com/Tencent-Hunyuan/SODA}} \\
}

\begin{document}
\maketitle
\makeatletter
\begingroup
  \renewcommand{\thefootnote}{}%
  \renewcommand{\@makefntext}[1]{\raggedright\noindent#1\par}%
  \footnotetext[0]{Qiuhong Ke is the corresponding author.
    \mbox{Research done} during Zinuo Li's internship \mbox{at Tencent Hunyuan.}}%
\endgroup
\makeatother

\begin{abstract}
    Video captioning directly reflects a Video Large Language Model’s perception, making caption fidelity a strong proxy for audio-visual grounding. Existing approaches often rely on costly annotations or stronger external teachers, limiting scalability. In this paper, we instead explore whether a VideoLLM can improve its own captions by exploiting structure already present within the model through self-supervision.
    Our exploration builds on two asymmetries we observe in current VideoLLMs. The first is a task asymmetry: captioning requires broad open-ended perception, whereas question answering narrows reasoning to specific dimensions. The second is an information asymmetry: the same model reasons more reliably with richer context or higher-fidelity audio-visual inputs.
    Motivated by these observations, we propose \emph{SODA} (\emph{\textbf{S}elf-Supervised \textbf{O}n-Policy \textbf{D}istillation via Dual-Stream \textbf{A}uditing}), a framework that enables a VideoLLM \rev{to} critique and refine its own captions. A student first produces an initial caption from a low-budget video view, followed by a dual-stream audit: a \emph{Caption QA Stream} identifies factual errors, and a \emph{Video QA Stream} detects omitted content. Conditioned on this Revision QA Bank, a teacher using a larger input budget rewrites the caption, and the refined outputs are distilled back into the student. Experiments suggest that, despite training only on the captioning objective, SODA improves caption fidelity, reduces hallucination, and remains competitive on general video understanding.
    \end{abstract}

\section{Introduction}
\label{sec:intro}

Improving captioning fidelity of Video Large Language Models (VideoLLMs) 
typically relies on either finer-grained multimodal annotations or stronger 
external teacher models~\citep{sharegpt4video2024,llavavideo2024}. Both are 
costly, hard to scale, and often brittle. Fine-grained video annotation is 
expensive and often emphasizes salient events rather than the fine-grained 
perceptual details that VideoLLMs frequently 
miss~\citep{vcapsbench2025,videohallucer2024}, leading to misaligned 
supervision. Meanwhile, teacher models trained through supervised fine-tuning or reinforcement learning methods such as RLHF and GRPO~\citep{ppo2017,grpo2024} typically provide supervision only through final responses, offering limited guidance for token-level supervision.

One approach to overcoming this limitation is On-Policy Distillation (OPD)~\citep{gkd2024,distillm2024}, which provides denser supervision by aligning student and teacher outputs at the token level along the student’s own trajectories. However, OPD still depends on stronger external teachers, whose training for omni-modal video understanding is itself expensive. We therefore ask a more modest question: can the model itself provide useful structure to guide its own captioning training, providing additional signal in a self-supervised manner? Rather than relying on stronger teachers as in OPD, we explore whether a low-resource, low-annotation approach that requires neither external teachers nor additional human labels can provide sufficient supervision on its own. Our approach is motivated by two asymmetries observed in current VideoLLMs as follows.

\begin{figure}[t]
\centering
\includegraphics[width=0.86\columnwidth]{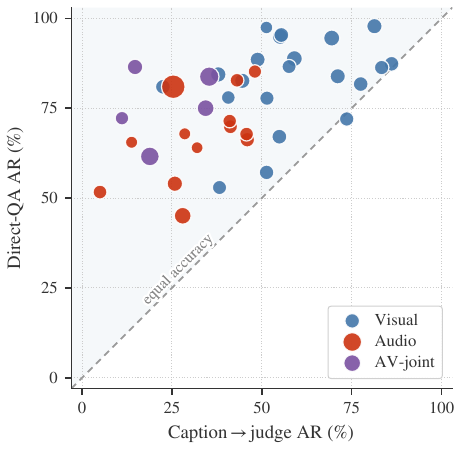}
\caption{\rev{QA accuracy versus caption accuracy across the dimensions in our caption-vs-QA probe. A point above the diagonal means the model answers the question more accurately from the video than from its own caption; color \rev{indicates the} modality and larger dots cover more questions. All dimensions but one lie above the line, so the model already knows more than it writes, and the gap is largest on audio and audio-visual content, where the caption often stays silent.}}
\label{fig:qa_vs_caption}
\end{figure}

The first is a \emph{task asymmetry}: \rev{the} captioning task is required to jointly cover all perceptual dimensions in a single open-ended generation, whereas question answering already narrows reasoning to specific dimensions. From the model's perspective, this difference manifests in attention: a question implicitly specifies what the model should attend to, whereas captioning requires the model to determine both where to attend and what to describe without an explicit query to guide attention~\citep{visatb2026}. This difficulty gap is what allows question answering to audit captioning: judging whether a caption is faithful does not require writing a better one; \rev{instead}, it only requires sampling questions from the caption and checking whether the answers generated from the video are consistent with the caption’s claims. Question answering can therefore provide supervision for captioning. \rev{Figure~\ref{fig:qa_vs_caption} and Table~\ref{tab:qa_vs_caption} show that this gap is real and large: holding the model and the questions fixed, answering a question directly from the video is more accurate than reading the same answer off the model's own caption, and the widest gaps fall on audio and audio-visual content, which is exactly where free-form captions tend to stay silent. Appendix~\ref{ssec:appendix-qa-vs-caption} gives the per-modality breakdown and the full protocol.}

\begin{table}[ht]
\centering
\small
\begin{tabular}{lccc}
\toprule
& AR$\uparrow$ & IR$\downarrow$ & CR$\uparrow$ \\
\midrule
caption$\to$judge & $43.5$ & $26.3$ & $59.1$ \\
direct-QA         & $\mathbf{77.3}$ & $\mathbf{14.2}$ & $\mathbf{90.1}$ \\
$\Delta$          & $+33.8$ & $-12.1$ & $+31.0$ \\
\bottomrule
\end{tabular}
\caption{\rev{Overall caption-vs-QA probe on the base model. Direct-QA answers each question from the video; caption$\to$judge reads the answer from the model's own caption. Per-modality numbers are in Appendix~\ref{ssec:appendix-qa-vs-caption}.}}
\label{tab:qa_vs_caption}
\floatvspace{-1em}
\end{table}

The second is an \emph{information asymmetry}: even with model parameters fixed, reasoning reliability improves when the model is provided with richer textual and multimodal information~\cite{klissarov2026improvinginteractiveincontextlearning,sang2026crispcompressedreasoningiterative}.
On the textual side, providing the model’s initial caption together with self-generated Revision QA pairs identifying potentially incorrect or incomplete parts of the caption improves reasoning performance, as feedback generated by the model itself has been shown to improve reasoning in prior work.
On the multimodal side, improving the fidelity of the same video and audio inputs through more sampled frames, higher spatial resolution, or longer audio context \rev{also} improves question-answering accuracy, with higher spatial resolution in particular consistently benefiting video understanding benchmarks. We refer to these as two \emph{information lifts}: both improve reasoning reliability without modifying model parameters, operating purely through richer inputs.

These two asymmetries motivate the design of our method \emph{Self-Supervised On-Policy Distillation via Dual-Stream Auditing} (SODA). The information asymmetry motivates a \rev{teacher-student} setup in which the same model operating under a richer input budget serves as the teacher, while the same model operating under a tighter input budget serves as the student. Because the teacher and student share parameters, the resulting supervision arises from differences in available information rather than model capacity. The task asymmetry chooses the curriculum: the model audits its own draft with two QA streams, one seeded from the caption (precision) and one from the audio-visual input (recall)~\citep{wang2023selfconsistency}; the teacher rewrites the initial caption conditioned on the resulting evidence. The student is then trained to match the teacher’s revised caption distribution by minimizing a token-level Kullback–Leibler divergence between their output distributions, providing dense per-token supervision beyond answer-level feedback. Because the student is trained under constrained input budgets, the distilled model tends to retain captioning quality and downstream performance even when the test-time input budget is reduced, improving token-budget efficiency for deployment.

We make two contributions.
\begin{itemize}
\setlength{\itemsep}{2pt}
\item We explore the feasibility of self-distillation \rev{for} video captioning without an external teacher or additional human labels by leveraging two key asymmetries in VideoLLMs: a task asymmetry that lets the model's own question answering audit its own captions, and an input-budget asymmetry that lets the same model reason more reliably under richer input.
\item We propose SODA, a self-supervised, self-evolving distillation framework that enables a VideoLLM to critique and refine its own captions. SODA uses dual-stream QA auditing to identify incorrect and missing content in the output caption, then distills the revised captions back into the model through dense token-level supervision.
\end{itemize}

\section{Related Works}
\label{sec:related}

SODA relates to three lines of work. \emph{Classical KD and on-policy distillation}~\citep{hinton2015distilling,kim2016seqkd,f-distill2024,minillm2024,gkd2024,distillm2024,gopd2026} relies on a teacher that is stronger than the student in parameters, data, or compute. \emph{Self-distillation and self-training}~\citep{bornagain2018,byot2019,meanteacher2017,noisystudent2020,stardb2023,rest2024,wang2023selfconsistency,videostar2024} typically separates teacher and student along a noise axis and supervises on the final answer. \emph{Hallucination mitigation for multimodal LLMs}~\citep{popebench2023,hallusionbench2024,videohallucer2024,vidhalluc2025,eventhallusion2025,avhbench2025,vcd2024,opera2024,rlhfv2024,povid2024,halva2024,lure2024,dpo2023,sdrl2026,videohallu2025,grpo2024,ppo2017,visualprm2026} relies on curated preference data, reward models, or an RL loop on top of answer tokens. SODA differs from all three: a weight-shared teacher whose edge is informational rather than capacity-based, a dense token-level KL on caption tokens that provides per-token soft targets, and no external annotation or reward signal. We expand each thread in Appendix~\ref{ssec:appendix-related-works}.

\section{Methodology}
\label{sec:method}

\subsection{Overview}
\label{sec:method:overview}
Inspired by the task asymmetry between captioning and question answering, and the information asymmetry induced by richer inputs, we propose \textit{\textbf{Self-Supervised On-Policy Distillation via Dual-Stream Auditing (SODA)}}, a self-distillation framework in which a single VideoLLM serves as both the student and teacher for its own captioning process. SODA retains the dense per-token supervision of on-policy distillation while replacing stronger external teachers with self-generated QA-based auditing signals, enabling training in low-resource settings without additional annotations. \textit{\textbf{The core idea of SODA is to create and then distill an information gap within the same model.}} The teacher branch receives both (1) richer audio-visual inputs and (2) QA-based feedback on the student’s initial caption, allowing it to generate more reliable revised captions. These revised outputs are then distilled back into the student, enabling the same model to retain strong performance even under constrained inference-time input budgets. 

\begin{figure*}[t]
\centering
\includegraphics[width=\textwidth]{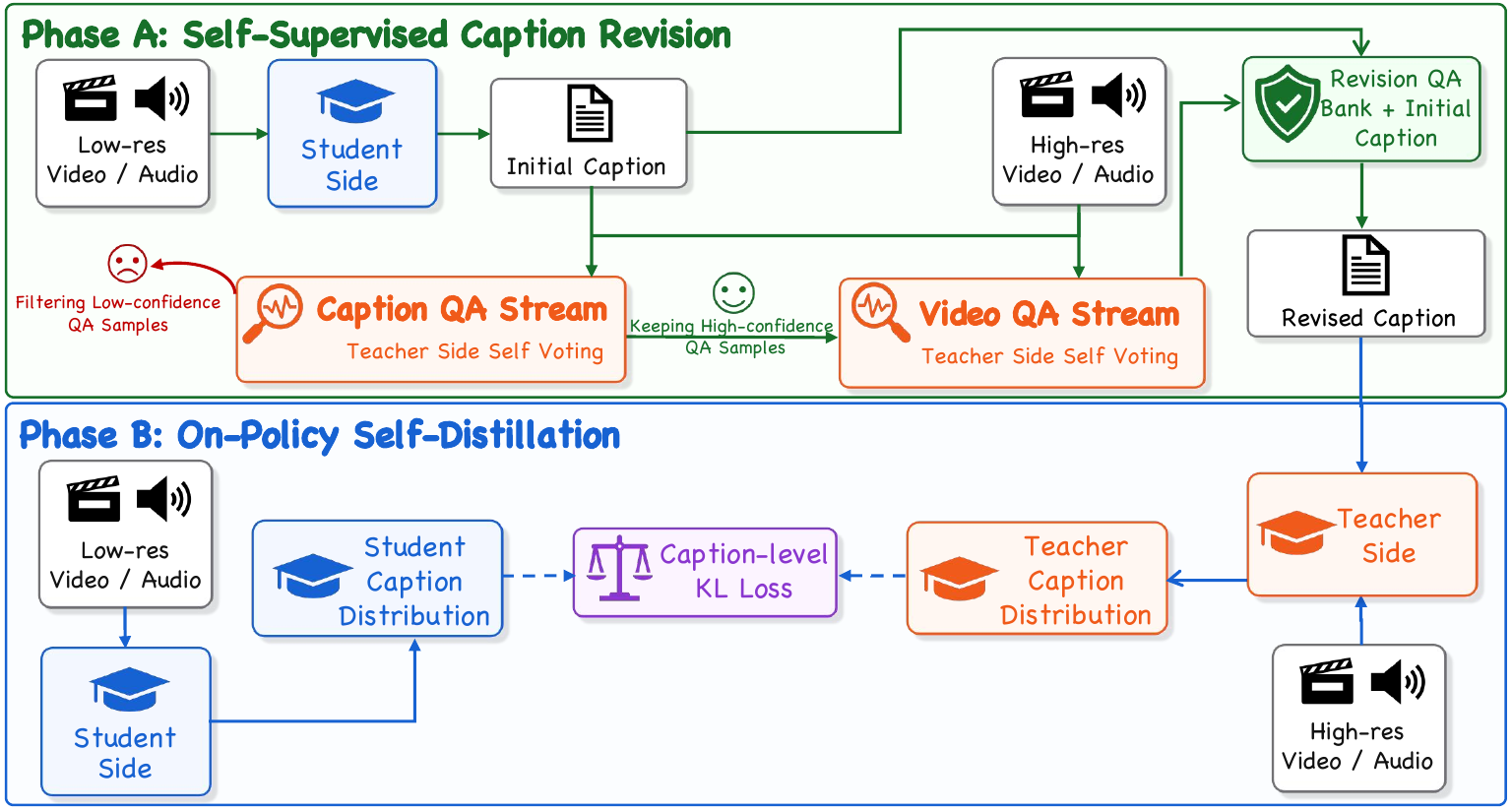}
\caption{Overview of SODA. Phase~A: the same model produces an initial caption $c_s$ from the low-budget view $V_{\text{lo}}$, audits it with the \emph{Caption QA Stream} (questions from $c_s$, catching commissions) and the \emph{Video QA Stream} (questions from the high-budget view $(V_{\text{hi}}, A_{\text{hi}})$, catching omissions); $K$-sample self-consistency voting yields the Revision QA Bank $\mathcal{E}_{\text{rqb}}$, and the teacher revises $c_s$ into a revised caption $c_r$. Phase~B: both roles are teacher-forced to $c_r$; we minimize a token-level KL between the student logits (conditioned on \rev{$V_{\text{lo}}$} and $\mathrm{prompt}_c$) and the teacher logits (conditioned on \rev{$(V_{\text{hi}}, A_{\text{hi}})$}, $c_s$, $\mathcal{E}_{\text{rqb}}$), with stop-gradient on the teacher side. The audit pipeline and the high-budget view are training-only.}
\label{fig:framework}
\floatvspace{-1em}
\end{figure*}

As illustrated in Figure~\ref{fig:framework}, training alternates two phases. In Phase A, the model first generates an initial caption, then audits it through two QA streams using richer audio-visual inputs, and finally revises the caption based on the resulting evidence. This produces a revised caption  $c_r$ that captures the additional information available to the teacher in textual form. In Phase B, this additional information is distilled back into the student. Both teacher and student are trained with the revised caption $c_r$ as the target, and training minimizes a token-level KL divergence between their caption distributions. Because $c_r$ is generated using richer inputs and QA-based auditing, the resulting dense per-token targets provide substantially more supervision than answer-level feedback alone. Importantly, both the auditing process and the richer input budget are used only during training, leaving inference-time computation unchanged.

\subsection{Phase A: Caption Revision}
\label{sec:method:phaseA}

Phase A enables the model to audit and revise its initial caption through QA-based self-reflection. The phase consists of three steps, each producing one component of a per-video tuple: (1) \emph{drafting} yields the student's initial caption $c_s$; (2) \emph{dual-stream auditing} yields the Revision QA Bank $\mathcal{E}_{\text{rqb}}$, a set of model-produced question and answer pairs that catch what $c_s$ gets wrong or misses; and \rev{(3)} \emph{revision} yields the revised caption $c_r$ obtained by rewriting $c_s$ based on $\mathcal{E}_{\text{rqb}}$. The resulting tuple $(c_s, \mathcal{E}_{\text{rqb}}, c_r)$ is cached and consumed by Phase~B.

\paragraph{Initial Caption.}
Let $M_\theta$ denote the VideoLLM with parameters $\theta$, and let $M_\theta(\cdot\mid\mathrm{ctx})$ \rev{denote} its autoregressive distribution over output tokens given a context. Given a video with visual and audio streams $(V, A)$ and a fixed captioning instruction $\mathrm{prompt}_c$, the student drafts an initial caption
\begin{equation}
c_s \sim M_\theta(\cdot \mid V, A, \mathrm{prompt}_c).
\label{eq:student_caption}
\end{equation}
This draft \rev{corresponds} to the student’s standard inference-time output and serves as the target of the subsequent auditing process.

\paragraph{Dual-Stream Auditing.}
A faithful caption should capture a broad range of perceptual aspects of a video, but a free-form draft $c_s$ often \rev{overlooks} important details. To explicitly define the scope of faithful captioning, we adopt a taxonomy of $D{=}\rev{29}$ audio-visual dimensions~\cite{vcapsbench2025} spanning visual content (e.g.\ subjects, actions, spatial layout, fine details, scene segments), camera and aesthetic (camera movement, shot type, \rev{depth of field}, color grading, visual style), on-screen text (native OCR, overlay text), audio (speech content, voice attributes, audio events, sound scene, audio-visual sync), and cross-modal checks (anti-hallucination, content type); the full list is given in Appendix~\ref{ssec:appendix-audit-taxonomy}. 

A typical $c_s$ omits many of these dimensions and may also contain uncertain or incorrect descriptions of the dimensions it does cover. The audit therefore serves two complementary purposes: verifying the claims made in $c_s$ and recovering content that $c_s$ omits or describes ambiguously.   A single QA stream cannot reliably accomplish both goals. Questions derived from the caption cannot probe content that the caption never mentions,  while questions derived from the video can recover missing content but are less effective at verifying specific claims already made in the caption. This motivates a \textbf{\textit{dual-stream auditing process}} in which caption-grounded and video-grounded QA evidence are combined and filtered through self-consistency voting.

To this end, we provide the auditing process with richer inputs than those used to generate the initial caption. Specifically, we construct a high-budget version $(V_{\text{hi}}, A_{\text{hi}})$ of the same video, with more sampled frames, higher resolution, and a longer audio window, and run the student on a tighter low-budget view $V_{\text{lo}}$. On top of this expanded view, every question $q$ is annotated with a dimension index $d\!\in\!\{1,\dots,D\}$ and a modality $m\!\in\!\{\text{visual},\text{audio}\}$. The two QA streams serve complementary roles and differ only in how their questions are generated:
\begin{itemize}\setlength{\itemsep}{1pt}
\item \textbf{Caption QA Stream} (\emph{verifying} the claimed dimensions). For each dimension, $M_\theta$ either generates a verification question grounded in $c_s$, or returns \texttt{SKIP\_NOT\_IN\_CAPTION} when $c_s$ does not address that dimension. The skipped dimensions are collected in $S_{\text{skip}}\!\subseteq\!\{1,\dots,D\}$, while the remaining questions form a candidate pool whose voted answers populate $\mathcal{E}^{\text{cap}}_{\text{rqb}}$.
\item \textbf{Video QA Stream} (\emph{\rev{recovering}} the missing dimensions). Let $U$ be the union of $S_{\text{skip}}$ and the dimensions whose Caption-stream answer is rejected or only weakly supported by the voting rule below (using a looser pivot threshold $\tau_{\text{piv}}\!\le\!\tau$), optionally capped at $D_{\max}$ entries per video. On the dimensions in $U$, $M_\theta$ generates fresh questions directly from $(V_{\text{hi}}, A_{\text{hi}})$, while avoiding previously used CaptionQA questions. The resulting answers form a second pool that populates $\mathcal{E}^{\text{vid}}_{\text{rqb}}$.
\end{itemize}

The two streams thus return question \emph{pools} but not yet trustworthy answers, since a single sample to an open-ended question may be unreliable. To obtain reliable revision signals, both streams use a shared \textbf{\emph{Self-Voting Mechanism}}. For each $q$, the teacher samples $K=5$ independent answers from the high-budget view,
\begin{equation}
a^{(k)} \sim M_\theta(\cdot \mid V_{\text{hi}}, A_{\text{hi}}, q),\quad k=1,\dots,K,
\label{eq:teacher_qa}
\end{equation}
under self-voting sampler $(T, p)$ with temperature $T=0.7$ and nucleus probability $p=0.9$. Before voting, answers are normalized so that semantically equivalent responses count as the same answer. Specifically, we strip trailing parenthetical justifications, lowercase and collapse whitespace, and map hedging phrases (e.g., ``not sure'', ``cannot tell'', ``unknown'') into a single ``uncertain'' bucket. The teacher then takes the majority answer $\hat a$ together with its agreement rate $\rho\!\in\![0,1]$, and retains $(q,\hat a)$ only if $\rho\!\ge\!\tau$ and $\hat a\!\neq\!\text{``uncertain''}$, the latter dropping questions on which the model itself declines to answer. The agreement $\rho$ thus serves as a self-calibrated reliability signal that filters out low-confidence or hedged questions before they enter the audit; the looser threshold $\tau_{\text{piv}}$ referenced above reuses this same signal to flag which Caption-stream dimensions need to be re-asked by the Video stream. Merging the survivors of both streams gives $\mathcal{E}_{\text{rqb}} = \mathcal{E}^{\text{cap}}_{\text{rqb}}\,\sqcup\,\mathcal{E}^{\text{vid}}_{\text{rqb}}$ (disjoint union after re-indexing). This model-generated QA set serves as revision evidence for the next stage.

\paragraph{Revision.}
The audit tells the teacher \emph{what} to fix; the revision step decides \emph{how}. We pack $c_s$ and $\mathcal{E}_{\text{rqb}}$ into a fixed revision-instruction template $\mathrm{prompt}_r(c_s, \mathcal{E}_{\text{rqb}})$ that asks the model to rewrite $c_s$ given $\mathcal{E}_{\text{rqb}}$, and the teacher rewrites the caption in a single greedy pass conditioned on the high-budget view,
\begin{equation}
\resizebox{\columnwidth}{!}{$\displaystyle
c_r = \arg\max\, M_\theta\!\left(\cdot \mid V_{\text{hi}}, A_{\text{hi}}, \mathrm{prompt}_r(c_s, \mathcal{E}_{\text{rqb}})\right).
$}
\label{eq:teacher_caption}
\end{equation}
The prompt treats each $\hat a_i$ as taking priority over $c_s$, and drops any audio claim whose entry has $\hat a_i\!=\!\texttt{none}$. Greedy decoding makes $c_r$ a stable target for Phase~B. We cache the tuple $(c_s, \mathcal{E}_{\text{rqb}}, c_r)$ for each training video and pass it on to Phase~B.

\subsection{Phase B: On-Policy Self-Distillation}
\label{sec:method:phaseB}

Phase~A produces $c_r$ as a richer text target, but writing better captions to a cache is not yet learning. Phase~B closes the loop by treating $c_r$ as the bridge through which the teacher's informational advantage is folded back into the student's parameters. For each cached tuple, we run two forward passes, both teacher-forced to $c_r$:
\begin{equation}
\resizebox{0.59\linewidth}{!}{$\displaystyle
\boldsymbol{\ell}_s = M_\theta\!\left(\rev{V_{\text{lo}}},\, \mathrm{prompt}_c;\ \mathrm{forced}{=}c_r\right)
$}
\label{eq:student_logits}
\end{equation}
\begin{equation}
\resizebox{0.85\linewidth}{!}{$\displaystyle
\boldsymbol{\ell}_t = M_\theta\!\left(\rev{V_{\text{hi}},\, A_{\text{hi}}},\, \mathrm{prompt}_r(c_s, \mathcal{E}_{\text{rqb}});\ \mathrm{forced}{=}c_r\right)
$}
\label{eq:teacher_logits}
\end{equation}
where $\boldsymbol{\ell}_s,\boldsymbol{\ell}_t\!\in\!\mathbb{R}^{|c_r|\times|\mathcal{V}|}$ are per-token logits over the model vocabulary $\mathcal{V}$ at every caption position, and $|c_r|$ denotes the token length of $c_r$. The captioning prompt $\mathrm{prompt}_c$ and the revision prompt $\mathrm{prompt}_r$ are the same fixed templates used in Phase~A; their full strings are given in Appendix~\ref{ssec:appendix-prompts}. \rev{The two paths differ by both information lifts: the teacher path conditions on $c_s$ and $\mathcal{E}_{\text{rqb}}$ and reads the high-budget view, while the student path has neither, so the KL transfers the text-side and the multimodal lift together; Section~\ref{sec:exp:ablation} separates them.} Only $\boldsymbol{\ell}_s$ carries gradients. The loss is the average token-level KL over the caption span,
\begin{equation}
\resizebox{\columnwidth}{!}{$\displaystyle
\mathcal{L}_{\text{SODA}}(\theta) = \frac{1}{|c_r|} \sum_{j=1}^{|c_r|} \mathrm{KL}\!\left(\, \mathrm{softmax}(\boldsymbol{\ell}_s^{(j)}) \,\Big\|\, \mathrm{sg}\!\left(\mathrm{softmax}(\boldsymbol{\ell}_t^{(j)})\right) \,\right),
$}
\label{eq:opd_loss}
\end{equation}
where $\mathrm{sg}(\cdot)$ is the stop-gradient operator. Intuitively, the student must recover the per-token distribution that the same model produces with $c_s$, $\mathcal{E}_{\text{rqb}}$\rev{, and the high-budget view}, while itself seeing \rev{none of them}; closing this gap forces the missing information to be encoded in the parameters rather than supplied \rev{at inference time}. We optionally replace KL with Jensen-Shannon (mixture coefficient $\beta\!\in\![0,1]$), which we find slightly more stable early in training.

\section{Experiments}
\label{sec:exp}

\subsection{Setup}
\label{sec:exp:setup}
We leave \rev{further} details in Appendix~\ref{ssec:appendix-experimental-setup}.
\paragraph{Backbone.}
We build SODA on Qwen2.5-Omni-7B~\citep{qwen25omni2025}, an open-source omni-modal model that processes video, audio, and text in a single transformer backbone. We keep the vision encoder and the audio tower frozen and insert LoRA adapters on the language backbone; the same checkpoint plays both student and teacher.

\paragraph{Training data.}
We use $2{,}000$ unlabeled videos \rev{sampled from Panda-70M~\citep{panda70m2024}, a publicly released collection of short clips}. Each video is processed once by Phase~A to produce a cached tuple $(c_s, \mathcal{E}_{\text{rqb}}, c_r)$, with no human annotation and \rev{no} overlap with any evaluation benchmark.

\paragraph{Caption-fidelity benchmarks.}
We evaluate captions with LLM judges. \emph{Omni-Cloze}~\citep{omnicloze2025} fills modality-specific blanks inside the predicted caption and reports per-modality and overall accuracy. \emph{UGC-VideoCap}~\citep{ugcvideocap2025} scores user-generated-content captions along visual, audio, and detail sub-dimensions. We supplement these benchmarks with \emph{VCapsBench-Omni} (Appendix~\ref{ssec:appendix-vcapsbench-omni}), which reports accuracy rate (AR$\uparrow$), inconsistency rate (IR$\downarrow$), and coverage rate (CR$\uparrow$) for visual, audio, and joint audio-visual questions.

\paragraph{General video understanding benchmarks.}
We use \emph{VideoHallucer}~\citep{videohallucer2024}, whose adversarial binary questions are paired into Basic (factual) and Hallucinated (counterfactual) on the same video, so that Overall requires both correct and is therefore sensitive to over-confident hallucination; and \emph{Video-MME~v2}~\citep{videomme2v2026}, which grades questions at three reasoning levels with a non-linear scoring scheme and draws most of its videos from after 2025. SODA does not directly optimize answer tokens on either, so improvements indicate caption-fidelity gains transferring to downstream general video understanding.

\subsection{Main Results: Caption Fidelity}
\label{sec:exp:caption}

Table~\ref{tab:caption} reports caption-fidelity results across the three benchmarks, comparing SODA against a panel of recent open-source VideoLLMs at a comparable parameter scale \rev{(7B to 8B)}. Rather than narrating per-cell deltas, we draw three observations.
\begin{itemize}\setlength{\itemsep}{2pt}
\item \textbf{Where the audit pushes is where the model improves.} The visible gains over Qwen2.5-Omni-7B fall on the audio and detail axes (Omni-Cloze audio, UGC-VideoCap audio and detail), the same dimensions on which the Revision QA Bank most often overturns or supplements the initial caption. On UGC-VideoCap, SODA scores lower than the base model on the visual sub-score.
\item \textbf{Recall, not precision, is the bottleneck the audit unblocks.} On VCapsBench-Omni the gain on coverage rate (CR) is far larger than on accuracy rate (AR), and inconsistency rate (IR) drifts up slightly. The IR drift is not supervision noise: omission, not commission, is the dominant failure mode of free-form captions (Appendix Figure~\ref{fig:audit_per_dim}), and once the video-grounded stream forces the model to address dimensions it would have skipped, some borderline calls surface. The audit turns a silent miss into an explicit, gradeable claim, which is the prerequisite for correcting it later.
\item \textbf{Per-token KL transfers granularity, not just coverage.} On UGC-VideoCap the largest lift is on the \emph{Detail} sub-score, which is hard to inflate by simply enumerating more entities. This is consistent with Phase~B teaching the student not only \emph{what} to mention but at \emph{what granularity}, since the teacher-forced $c_r$ encodes both choices jointly token by token.
\end{itemize}

\begin{table*}[t]
\centering
\small
\resizebox{\textwidth}{!}{%
\begin{tabular}{lccccccccccccc}
\toprule
& \multicolumn{4}{c}{Omni-Cloze\rev{$^{\dagger}$}} & \multicolumn{3}{c}{VCapsBench-Omni} & \multicolumn{4}{c}{UGC-VideoCap (Final Caption)\rev{$^{\dagger}$}} \\
\cmidrule(lr){2-5} \cmidrule(lr){6-8} \cmidrule(lr){9-12}
Method & Vis$\uparrow$ & Aud$\uparrow$ & AV$\uparrow$ & Tot$\uparrow$ & AR$\uparrow$ & IR$\downarrow$ & CR$\uparrow$ & Aud$\uparrow$ & Vis$\uparrow$ & Det$\uparrow$ & Avg$\uparrow$ \\
\midrule
\multicolumn{12}{l}{\emph{Proprietary Omni-Modal Models}} \\
Gemini 2.0 Flash~\citep{gemini20flash2025}   & 32.3 & 31.7 & 40.1 & 33.2 & - & - & - & - & - & - & - \\
Gemini 2.5 Flash~\citep{gemini25flash2025}   & 24.4 & 36.2 & 41.4 & 33.0 & - & - & - & 74.2 & 78.8 & 77.2 & 76.7 \\
Gemini 2.5 Pro~\citep{gemini25pro2025}       & 40.8 & 44.1 & 52.8 & 43.6 & 62.1 & 15.4 & 73.5 & 70.8 & 75.8 & 74.8 & 73.8 \\
\midrule
\multicolumn{12}{l}{\emph{Open-Source Omni-Modal LLMs}} \\
video-SALMONN-13B~\citep{videosalmonn2024}   & 3.5  & 2.3  & 4.6  & 3.3  & - & - & - & - & - & - & - \\
VideoLLaMA2-7B~\citep{videollama22024}       & 7.6  & 4.3  & 8.7  & 6.6  & - & - & - & - & - & - & - \\
Gemma-3n-E2B-it~\citep{gemma3n2025}          & - & - & - & - & 46.0 & \second{29.3} & 67.4 & 52.0 & 52.4 & 49.6 & 51.3 \\
Gemma-3n-E4B-it~\citep{gemma3n2025}          & - & - & - & - & \best{47.8} & 29.7 & \best{69.4} & 54.2 & 54.6 & 51.2 & 53.3 \\
Qwen2.5-Omni-3B~\citep{qwen25omni2025}       & - & - & - & - & 37.6 & \best{28.5} & 52.6 & 48.2 & 55.6 & 52.6 & 52.2 \\
MiniCPM-o-2.6-8B~\citep{minicpmo2024}        & - & - & - & - & 30.6 & 29.7 & 41.7 & 46.0 & \best{59.7} & \second{62.4} & 56.3 \\
\rev{video-SALMONN 2+}~\rev{\citep{tang2025video}} & \rev{-} & \rev{-} & \rev{-} & \rev{\best{28.3}} & \rev{37.0} & \rev{-} & \rev{47.2} & \rev{-} & \rev{-} & \rev{-} & \rev{\best{68.1}} \\
\midrule
Qwen2.5-Omni-7B (base)~\citep{qwen25omni2025} & \second{18.3} & \second{14.1} & \second{21.9} & \rev{16.6} & 44.5 & 29.7 & 60.7 & \second{57.6} & \second{59.4} & 57.0 & \rev{58.0} \\
\,\,+ SODA (ours)                              & \best{27.1} & \best{21.9} & \best{24.8} & \rev{\second{25.0}} & \second{47.7} & 30.3 & \second{68.4} & \best{73.1} & 44.2 & \best{73.3} & \rev{\second{63.5}} \\
\bottomrule
\end{tabular}%
}
\caption{Caption-fidelity results on Omni-Cloze (Visual / Audio / Audio-Visual / Total accuracy), VCapsBench-Omni (AR / IR / CR), and UGC-VideoCap (Final Caption sub-scores: Audio / Visual / Detail / Average). ``-'' marks \rev{that the metrics} are not available on the corresponding models. Per column, the best open-source result is in \best{blue bold} and the second-best is \second{underlined}. \rev{$^{\dagger}$~quoted from official papers for all rows but ours; VCapsBench-Omni is run by us for every model.} The UGC-VideoCap scores for SODA aggregate the 556 valid GPT-4o judgments out of 4,800 questions.}
\label{tab:caption}
\floatvspace{-1em}
\end{table*}

\subsection{Main Results: General Video Understanding Transfer}
\label{sec:exp:mcq}

SODA's training loss touches only the captioning span; it never sees an MCQ answer token. Table~\ref{tab:mcq} therefore reports the more pointed question of whether captioning-only self-supervision \emph{costs} general video understanding. We highlight three observations.
\begin{itemize}\setlength{\itemsep}{2pt}
\item \textbf{Captioning-only training even slightly improves general ability.} On VideoHallucer the SODA-trained student improves on all three slices (Basic, Hallucinated, and the stricter Overall that requires both correct on the same video); on Video-MME~v2 it scores slightly above the base model at Levels~2 and~3 and slightly below it at Level~1. This is the opposite of the side effect typically observed when a narrow objective is sharpened by SFT, where probes outside the loss often \emph{regress} as the price of the in-task gain.
\item \textbf{On-policy targets keep the supervision inside the model's own support.} We attribute this benign side effect, tentatively, to the on-policy and self-supervised nature of the target. The revised caption $c_r$ is greedily decoded by the same model from inputs it has already encoded, and Phase~B's teacher-forced KL only pulls the student toward a distribution that already lies within its own support: there is no foreign distribution to be \emph{covered}, only a redistribution of mass within \rev{one that} the model already represents. Conventional SFT on captions written by a different annotator, by contrast, asks the model to assign mass to sequences far from its prior, and the optimizer can satisfy that demand by overwriting capabilities that were not part of the loss. We do not claim a causal mechanism on the basis of two probes; what the experiments show is that the side effect of SODA on the general-purpose probes we measured is, at minimum, not harmful.
\item \textbf{Gains concentrate on the slices where audio-visual fidelity matters most.} The VideoHallucer gain is \rev{larger} on the \emph{Hallucinated} slice (which penalizes over-claiming content not supported by the input) \rev{than on \emph{Basic}, and largest on \emph{Overall}}; on Video-MME~v2 the gains concentrate on Levels~2 and~3, the harder reasoning tiers. Both patterns are consistent with the audit specifically removing unsupported claims from $c_r$ during training, so what transfers to MCQ time is not only a richer caption distribution but a calibrated reluctance to assert content the audio-visual evidence does not back.
\end{itemize}

\begin{table*}[t]
\centering
\small
\resizebox{\textwidth}{!}{%
\begin{tabular}{lcccccc}
\toprule
& \multicolumn{3}{c}{VideoHallucer\rev{$^{\dagger}$}} & \multicolumn{3}{c}{Video-MME~v2 (w/o subtitles)\rev{$^{\dagger}$}} \\
\cmidrule(lr){2-4} \cmidrule(lr){5-7}
Method & Basic$\uparrow$ & Hallu.$\uparrow$ & Overall$\uparrow$ & Level 1 & Level 2 & Level 3 \\
\midrule
\multicolumn{7}{l}{\emph{Proprietary Omni-Modal Models}} \\
GPT4-o~\cite{gpt4o2024}                  & 75.1 & 74.2 & 53.3 & - & - & - \\
\midrule
\multicolumn{7}{l}{\emph{Open-Source VideoLLMs}} \\
Video-LLaMA2-7B~\citep{videollama22024}                  & \second{90.9} & 12.7 & 10.0 & - & - & - \\
Video-LLaVA-7B~\citep{videollava2024}                    & \best{95.1} & 20.3 & 17.8 & - & - & - \\
PLLaVA-7B~\citep{pllava2024}                             & 75.1 & 55.5 & 38.1 & - & - & - \\
LLaVA-NeXT-Video-DPO-7B~\citep{llavanextvideodpo2024}    & 62.5 & \second{60.9} & 32.0 & - & - & - \\
LLaVA-Video-7B-Qwen2~\citep{llavavideo2024}              & - & - & - & 12.7 & 5.6  & 5.1  \\
VideoLLaMA3-7B~\citep{videollama32025}                   & - & - & - & 15.8 & 9.2  & 9.1  \\
Qwen2.5-VL-7B-Instruct~\citep{qwen25vl2025}              & - & - & - & 15.3 & 13.7 & 12.6 \\
InternVL3.5-8B-Instruct~\citep{internvl35}               & - & - & - & 13.5 & 10.8 & 10.7 \\
\midrule
Qwen2.5-Omni-7B (base)~\citep{qwen25omni2025}            & 73.5 & 58.1 & \second{44.8} & \best{32.6} & \second{17.8} & \second{19.1} \\
\,\,+ SODA (ours)                                          & 75.9 & \best{62.4} & \best{51.5} & \second{32.1} & \best{18.7} & \best{19.8} \\
\bottomrule
\end{tabular}%
}
\caption{General video understanding accuracy (\%). VideoHallucer reports Basic (factual), Hallucinated (counterfactual), and Overall accuracy where Overall requires both to be correct on the same video; ``-'' marks \rev{that the metrics} are not available on the corresponding models. Per column, the best open-source result is in \best{blue bold} and the second-best is \second{underlined}. \rev{$^{\dagger}$~quoted from official papers for all rows but ours.}}
\label{tab:mcq}
\floatvspace{-1em}
\end{table*}

\subsection{Ablations}
\label{sec:exp:ablation}

We report four ablations that target the key design choices of SODA, summarized in Table~\ref{tab:ablation}. The first row repeats the base Qwen2.5-Omni-7B as the Round~0 reference. \rev{Each row turns off or changes one part of the recipe. We start before training: \emph{Revised Caption Only ($c_r$)} is the caption Phase~A writes on the base model, with Phase~B never run, so it shows what the audit alone can do. The next two rows take away one of the teacher's two advantages each. \emph{w/o Extra Vid/Aud Info} keeps the full audit but drops the teacher to the student's low video and audio budget, so only the input-budget lift is gone. \emph{w/o Audit} does the reverse: the teacher keeps the high budget but writes one caption straight away, with no questions asked. Having isolated what the teacher sees, the next two rows ask when its weights are read. \emph{EMA update} takes a moving average of the student between rounds and \emph{Joint update} reads the current student weights under a no-grad forward pass, while the audit, the voting, the loss, and the input budgets all stay as they are in the default frozen teacher. \emph{Single-Stream} then narrows the audit itself, keeping only the questions seeded from the caption and dropping the ones seeded from the video. Finally, \emph{SODA Round 1} and \emph{Round 3} run the same recipe for fewer or more rounds, and the last row is the full recipe we use everywhere else in the paper.}

\paragraph{Phase~A produces the target, Phase~B is what trains the model.}
The \emph{Revised Caption Only ($c_r$)} row evaluates the revised caption $c_r$ produced by Phase~A on the base model, with no Phase~B training applied. It already moves the base on Omni-Cloze visual and total, confirming that the audit-and-revise pipeline produces a meaningfully better caption than the model's own draft. The full SODA recipe then gains a further, comparable amount on top, particularly on audio sub-scores and on coverage rate. The gap between \emph{Revised Caption Only ($c_r$)} and full SODA is the share of the improvement that is \emph{not} accessible by prompting harder at inference: Phase~B's per-token KL turns the one-shot textual lift into a property of the parameters, which the student carries to videos it has never seen revisions for and to inputs where the audit pipeline is unavailable. As the \emph{\#Input Tok.} column shows, Phase~B also compresses the Phase~A inference cost by ${\sim}8\times$; Appendix Figure~\ref{fig:cost_vs_quality} plots the Pareto picture explicitly.

\begin{table*}[t]
\centering
\small
\resizebox{\textwidth}{!}{%
\begin{tabular}{lccccccccccc}
\toprule
& \multicolumn{4}{c}{Omni-Cloze} & \multicolumn{3}{c}{VCapsBench-Omni} & \multicolumn{3}{c}{Video-MME~v2 (w/o sub)} & \\
\cmidrule(lr){2-5} \cmidrule(lr){6-8} \cmidrule(lr){9-11}
Setting & Vis$\uparrow$ & Aud$\uparrow$ & AV$\uparrow$ & Tot$\uparrow$ & AR$\uparrow$ & IR$\downarrow$ & CR$\uparrow$ & L1 & L2 & L3 & \#Input Tok.$\downarrow$ \\
\midrule
Qwen2.5-Omni-7B~\citep{qwen25omni2025} & 18.3 & 14.1 & 21.9 & 16.6 & 44.5 & \rev{29.7} & 60.7 & 32.6 & 17.8 & 19.1 & 830 \\
\midrule
Revised Caption Only ($c_r$)           & \best{27.3} & 13.0 & 22.3 & 21.8 & 45.8 & \best{26.6} & 62.4 & - & - & - & 6{,}936 \\
w/o Extra Vid/Aud Info  & 26.5 & 20.0 & 24.4 & 23.9 & 46.8 & 29.1 & 67.8 & 31.3 & 17.2 & 19.1 & 830 \\
w/o Audit               & 20.3 & 8.8 & 16.9 & 15.9 & 45.4 & 29.2 & \rev{63.8} & 30.7 & 16.5 & 18.6 & 830 \\
EMA update                                  & 26.7 & 22.2 & \best{25.0} & \second{24.9} & \second{47.6} & 30.1 & \second{68.2} & 31.3 & \second{18.6} & 19.0 & 830 \\
Joint update                                & 25.9 & 22.5 & 22.8 & 24.3 & 47.2 & 29.1 & 66.5 & 31.6 & \best{18.7} & \second{19.6} & 830 \\
Single-Stream      & 25.0 & \second{23.5} & 24.4 & 24.6 & 46.2 & \second{28.1} & 65.6 & 30.2 & 17.8 & 18.5 & 830 \\
SODA Round 1            & 26.7 & 19.2 & 22.3 & 23.6 & \second{47.6} & 29.4 & 67.8 & \second{32.0} & 17.4 & 19.0 & 830 \\
SODA Round 3             & 22.4 & \best{24.9} & 20.9 & 23.0 & 45.8 & 29.9 & 65.3 & 30.5 & \best{18.7} & \best{19.8} & 830 \\
\textbf{SODA (Frozen Teacher+Dual Stream+Round2)} & \second{27.1} & 21.9 & \second{24.8} & \best{25.0} & \best{47.7} & 30.3 & \best{68.4} & \best{32.1} & \best{18.7} & \best{19.8} & 830 \\
\bottomrule
\end{tabular}%
}
\caption{Ablation studies\rev{; unless the row name says otherwise, every variant is trained for 2 rounds}. The variants are defined in Section~\ref{sec:exp:ablation}. The first row repeats the base Qwen2.5-Omni-7B numbers from Tables~\ref{tab:caption} and~\ref{tab:mcq}. \emph{\#Input Tok.}: average text tokens consumed at \emph{inference} time per video to produce the evaluated caption (excluding video frame and audio tokens). Per column, the best ablation variant is in \best{blue bold} and the second-best is \second{underlined} (the reference row is excluded from the comparison).}
\label{tab:ablation}
\end{table*}

\paragraph{The input gap is the signal, a moving teacher doesn't mean better.}
Because student and teacher share $\theta$, when the teacher's weights are read is a genuine design choice. We compare (i)~\emph{frozen teacher} (our default), with the checkpoint fixed at pre-Phase-B; (ii)~\emph{EMA update}, refreshed as an exponential moving average of the student between rounds; and (iii)~\emph{joint update}, where the teacher branch always reads the current student weights under a no-grad forward pass. The teacher already differs from the student through access to $c_s$, $\mathcal{E}_{\text{rqb}}$, and a higher-fidelity audio-visual view, so stacking a moving-weight gap on top of an already informative input gap mostly introduces optimization noise without changing what is being taught. Freezing the teacher is both the simplest and the most stable choice.

\paragraph{Each stream owns one side, together they cover both.}
The \emph{Single-Stream} ablation drops the video-grounded pivot stream, leaving the audit with only questions seeded from $c_s$. On VCapsBench-Omni this has the expected shape: IR \emph{improves} because the student is only ever asked to defend claims it actually made, while CR \emph{drops} because dimensions the caption never raised cannot be questioned at all. The dual-stream default trades a small amount of precision for a larger amount of recall, which is the right side of the curve on benchmarks where omission, not commission, is the dominant failure mode. Caption-grounded for \emph{check}, video-grounded for \emph{recover}: neither subsumes the other.

\paragraph{Self-evolution helps, then plateaus.}
Treating the base model as Round~0 and refreshing Phase~A on the round-$r$ student each iteration, the procedure improves through Round~2 and then mildly regresses at Round~3 on Omni-Cloze \rev{(Appendix Figure~\ref{fig:rounds_plateau}), while the Video-MME~v2 columns of Table~\ref{tab:ablation} stay roughly flat}. We attribute this pattern to a gradual saturation of the audit: as the student improves, we expect the pool of dimensions on which the audit can still extract a confident correction to shrink, so each additional refresh both adds less new signal and risks reinforcing the student's residual biases against itself. We take Round~2 as the default operating point.

\subsection{\rev{Generalization to Other Base Models}}
\label{sec:exp:generalization}

\rev{We also study the generalization ability of SODA by repeating the whole pipeline on a smaller model, Qwen2.5-Omni-3B, and on a different model family, MiniCPM-o-2.6.}

\begin{table}[ht]
\centering
\small
\resizebox{\columnwidth}{!}{%
\begin{tabular}{llccc}
\toprule
Backbone & Benchmark & base & +SODA & $\Delta$ \\
\midrule
\multirow{4}{*}{Qwen2.5-Omni-3B}
  & Omni-Cloze (Tot$\uparrow$)   & 17.24 & 19.61 & +2.37 \\
  & VCapsBench-Omni (AR$\uparrow$) & 37.6  & 42.51 & +4.91 \\
  & VCapsBench-Omni (CR$\uparrow$) & 52.6  & 60.87 & +8.27 \\
  & Video-MME~v2 (Acc$\uparrow$)  & 25.56 & 25.83 & +0.27 \\
\midrule
\multirow{4}{*}{MiniCPM-o-2.6}
  & Omni-Cloze (Tot$\uparrow$)   & 22.79 & 23.58 & +0.79 \\
  & VCapsBench-Omni (AR$\uparrow$) & 30.6  & 33.81 & +3.21 \\
  & VCapsBench-Omni (CR$\uparrow$) & 41.7  & 45.64 & +3.94 \\
  & Video-MME~v2 (Acc$\uparrow$)  & 23.09 & 23.75 & +0.66 \\
\bottomrule
\end{tabular}%
}
\caption{Generalization studies of SODA \rev{applied} to different omni-modal backbones under the same training protocol. Video-MME~v2 is reported here as overall accuracy, while Table~\ref{tab:mcq} reports the benchmark's per-level scores.}
\label{tab:backbones}
\end{table}

\rev{Table~\ref{tab:backbones} shows that both additional backbones move in the same direction as the 7B result, so the gain does not come from one particular checkpoint. It is bigger on the Qwen2.5-Omni family and smaller on MiniCPM-o-2.6, which fits how SODA gets its signal: a model that follows instructions less well writes a noisier Revision QA Bank, and a noisier bank gives a weaker revision target. Even then, coverage rate still goes up, so a weaker base model limits the method rather than breaking it. General video understanding again barely moves, which is the same picture we saw in Table~\ref{tab:mcq}.}
\section{Conclusion}
\label{sec:conclusion}

In this paper, \rev{we start from} two structural asymmetries \rev{that} already live inside current VideoLLMs: captioning tends to be harder than answering questions sampled from a caption, and the same model reasons more reliably with a richer input budget. We then turn these into a self-distillation loop. The same model plays a high-budget teacher and a low-budget student; a dual-stream audit, filtered by self-consistency voting, builds a Revision QA Bank $\mathcal{E}_{\text{rqb}}$ that flags which caption dimensions need fixing; a token-level KL on the revised caption then transfers the teacher's evidence into the student's parameters without any human labels. SODA improves caption fidelity across three benchmarks and raises VideoHallucer scores. Its Video-MME~v2 scores remain close to those of the base model. \rev{The gains also extend to two additional backbones. Because the audit and the high-budget view are training-only, the distilled student keeps the base model's inference cost.} \rev{Two insights generalize} beyond captioning: \rev{a} ``stronger teacher'' can simply be the same model under a richer input budget, \rev{and} which tokens carry the supervision matters as much as how much supervision is provided.

\section*{Limitations}
\label{sec:limitations}
We believe a systematic scaling-law study along both \rev{the model and data} fronts is an important step in the future. Beyond scale, the self-supervised nature of SODA comes with a natural trade-off. We hope to scale up this approach in the future by injecting a small pool of ground-truth in a semi-supervised setup, or by injecting auxiliary meta-information beyond the raw audio-visual stream.

\clearpage
\section*{Acknowledgements}
This research was supported by the Australian Government through the Australian Research Council’s Discovery Projects funding scheme (project numbers DP260100218 and DP260101891). Dr Qiuhong Ke is the recipient of an Australian Research Council Discovery Early Career Researcher Award (project number DE250100030) funded by the Australian Government.

\bibliography{custom}
\appendix
\section{Appendix}
\label{sec:appendix}

\subsection{Experimental Setup Details}
\label{ssec:appendix-experimental-setup}

This subsection collects the hyperparameters and implementation details deferred from Section~\ref{sec:exp:setup}.

\paragraph{Information lifts.}
The student view $V_{\text{lo}}$ uses $F_s\!\le\!100$ frames sampled at $1$~fps from a per-frame pixel budget $R_s\!=\!151{,}200$ and a $T_s\!=\!120$~s audio window at $16$~kHz mono; the teacher view $(V_{\text{hi}}, A_{\text{hi}})$ uses $F_t\!\le\!100$ frames at $2$~fps from a per-frame pixel budget $R_t\!=\!453{,}600$ ($3{\times}$ the student) and a $T_t\!=\!240$~s audio window ($2{\times}$ the student), so the teacher consistently observes a richer audio-visual budget. Phase~A captions are sampled with temperature $0.7$ inside a length window of $[80, 4096]$ tokens, and the QA generator emits exactly $28$ questions per video. Self-Voting uses $K\!=\!5$ samples per question under temperature $T\!=\!0.7$ and nucleus $p\!=\!0.9$, with agreement thresholds $\tau\!=\!0.6$ for Stream-1 retention and $\tau_{\text{piv}}\!=\!0.6$ for Stream-2 pivot triggering, capped at $D_{\max}\!=\!8$ pivot dimensions per video. Phase~A is refreshed every $R$ optimization steps.

\paragraph{Implementation details.}
We instantiate SODA on Qwen2.5-Omni-7B. Phase~A runs in inference mode with data-parallel sharding across the available accelerators; Phase~B uses standard distributed data-parallel training. Phase~B trains LoRA adapters of rank~$64$ with $\alpha\!=\!128$ on the $\{q,k,v,o\}_{\text{proj}}$ matrices of the language backbone, keeping the visual and audio encoders frozen. Optimization uses AdamW with a constant learning rate $5{\times}10^{-5}$ and $5\%$ linear warm-up, per-device batch size $2$ with gradient accumulation $1$, maximum sequence length $32{,}768$ tokens, and a maximum completion length of $4{,}096$ tokens. The loss is the KL form in Eq.~\ref{eq:opd_loss} with softmax temperature $1.0$; we optionally replace it with Jensen-Shannon (mixture coefficient $\beta\!=\!0.7$). Training runs for \rev{$2$} on-policy rounds; each round consists of one Phase~A cache regeneration followed by one Phase~B epoch over a $2{,}000$-video \rev{subset sampled from Panda-70M}, and successive rounds warm-start from the previous round's merged adapter.

\subsection{Asking the Question Directly: a QA-vs-Caption Probe on the Base Model}
\label{ssec:appendix-qa-vs-caption}

The dual-stream audit in Phase~A is built on the claim that, for the same omni-modal model, \emph{answering a yes/no question grounded in audio-visual evidence is a more accurate task than writing a free-form caption and reading the answer back out of the caption} (Section~\ref{sec:method:overview}, ``task asymmetry between captioning and question answering''). This subsection reports the simplest possible direct check of that claim on the base model SODA is built on.

\paragraph{Setup.}
We hold the model fixed at the base Qwen2.5-Omni-7B and run two strictly parallel pipelines on the same 100-video subset of VCapsBench-Omni (seed-42 random draw), using the yes/no QAs in the intersection across visual, audio, and AV-joint tracks.
\begin{itemize}\setlength{\itemsep}{1pt}
\item \textbf{Caption$\to$judge.} The same protocol as the main-table evaluation (Section~\ref{sec:exp:caption}). The model emits one omni caption per video; Gemini-2.5-Pro then reads each GT QA off the caption text alone and returns \{\textsc{Yes}, \textsc{No}, \textsc{Unanswerable}\}.
\item \textbf{Direct-QA.} For each GT QA, the same model is shown the video and audio with a one-line yes/no/unsure instruction and is asked to commit to a verdict directly. No external judge is involved; the model's own answer is graded against GT.
\end{itemize}
The two pathways score with the same AR / IR / CR formula on the same (video, question) pairs, so the \rev{comparison} is apples-to-apples. The judge in the caption pathway is strictly stronger than the model, which biases the comparison \emph{against} our hypothesis.

\paragraph{Results.}
Table~\ref{tab:qa_vs_caption} in the main paper reports the overall numbers, and Table~\ref{tab:qa_vs_caption_detail} together with Figure~\ref{fig:qa_vs_caption} give the per-modality view. The direct-QA pathway is more accurate by a large margin on every track, gaining $+33.8$ AR overall and improving IR and CR as well. The per-modality breakdown is the predicted shape: the gap is smallest on visual, where captions tend to enumerate scene content anyway, and widest on audio and AV-joint, where free-form omni captions routinely omit the dimension entirely and the judge has no choice but to mark the QA \textsc{Unanswerable}. CR makes this concrete: on AV-joint the caption pathway can only attempt about two out of five QAs, while direct-QA commits to nearly all of them. Figure~\ref{fig:qa_vs_caption} confirms this at finer resolution: all but one of the VCapsBench-Omni dimensions that move between the two pipelines move upward, and the largest jumps concentrate on the audio (red) and AV-joint (purple) clouds in the upper-left.

\paragraph{Per-QA agreement.}
The same conclusion sharpens at the per-QA level. On the intersected QAs, the direct-QA pathway recovers several times more QAs than it loses. The overwhelming majority of those direct-QA-only wins are cases where the caption pathway returned \textsc{Unanswerable} (the caption was silent on that dimension) and the model nonetheless committed to the correct yes/no when asked directly. These are exactly the items that the Video QA Stream in Phase~A is designed to surface and that the caption-only audit would miss by construction.

\paragraph{Why this motivates dual-stream auditing.}
The QA-vs-caption gap above is what the audit converts into supervision. The Caption QA Stream (Section~\ref{sec:method:phaseA}) targets the cases where the caption \emph{does} commit to a claim but the claim is wrong, which is the IR row of Table~\ref{tab:qa_vs_caption} and corresponds to the \emph{check} job. The Video QA Stream targets the large share of probes that the caption stays silent on ($1{-}\mathrm{CR}$, overall column) while the model can answer them directly, which is the \emph{recover} job. Both classes are large, both are concentrated on the audio and AV-joint sides, and both vanish from any supervision signal that lives on the caption text alone. Folding both into $c_r$ and then distilling $c_r$ back into the student in Phase~B is the mechanism by which SODA turns ``the model knows more than its caption admits'' into a parameter-level gain.

\begin{table}[t]
\centering\small
\begin{tabular}{lrrr}
\toprule
& AR$\uparrow$ & IR$\downarrow$ & CR$\uparrow$ \\
\midrule
\multicolumn{4}{l}{\emph{Visual} ($n{=}1{,}716$)} \\
\quad caption$\to$judge & $58.1$ & $17.1$ & $70.1$ \\
\quad direct-QA         & $\mathbf{83.2}$ & $\mathbf{12.8}$ & $\mathbf{95.4}$ \\
\quad $\Delta$          & $+25.1$ & $-4.3$ & $+25.3$ \\
\midrule
\multicolumn{4}{l}{\emph{Audio} ($n{=}1{,}059$)} \\
\quad caption$\to$judge & $31.0$ & $42.9$ & $54.2$ \\
\quad direct-QA         & $\mathbf{68.8}$ & $\mathbf{16.6}$ & $\mathbf{82.5}$ \\
\quad $\Delta$          & $+37.9$ & $-26.3$ & $+28.3$ \\
\midrule
\multicolumn{4}{l}{\emph{AV-joint} ($n{=}652$)} \\
\quad caption$\to$judge & $25.7$ & $32.8$ & $38.2$ \\
\quad direct-QA         & $\mathbf{75.5}$ & $\mathbf{14.7}$ & $\mathbf{88.5}$ \\
\quad $\Delta$          & $+49.8$ & $-18.1$ & $+50.3$ \\
\bottomrule
\end{tabular}
\caption{Per-modality Caption$\to$judge vs.\ Direct-QA on the same 100-video probe as Table~\ref{tab:qa_vs_caption}. The judge for the caption pathway is Gemini-2.5-Pro. Direct-QA is graded against GT without any judge. $\Delta$ rows are direct-QA minus caption$\to$judge.}
\label{tab:qa_vs_caption_detail}
\end{table}

\subsection{Supplementary Figures for Ablations}
\label{ssec:appendix-ablation-figures}

This subsection collects two visualizations of the ablation results in Table~\ref{tab:ablation}, both referenced from Section~\ref{sec:exp:ablation}.

\begin{figure}[t]
\centering
\includegraphics[width=\columnwidth]{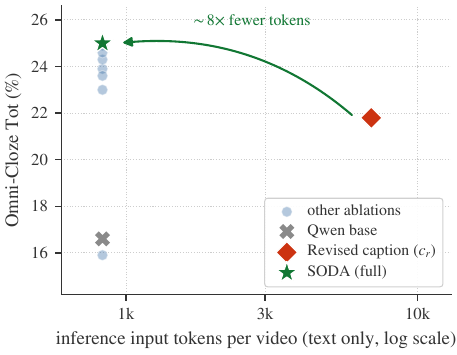}
\caption{Inference cost vs.\ caption-fidelity quality across the ablation variants of Table~\ref{tab:ablation}. The $x$-axis is per-video text input tokens at inference time (log scale; video frame and audio tokens excluded); the $y$-axis is Omni-Cloze Tot.\ accuracy. \emph{Revised Caption Only ($c_r$)} sits at ${\sim}6.9$k tokens because every inference must rerun the full Phase~A pipeline, while all Phase~B-trained variants run the student on the standard caption prompt alone ($830$ tokens) and the full SODA recipe dominates on both axes simultaneously.}
\label{fig:cost_vs_quality}
\end{figure}

\begin{figure}[t]
\centering
\includegraphics[width=\columnwidth]{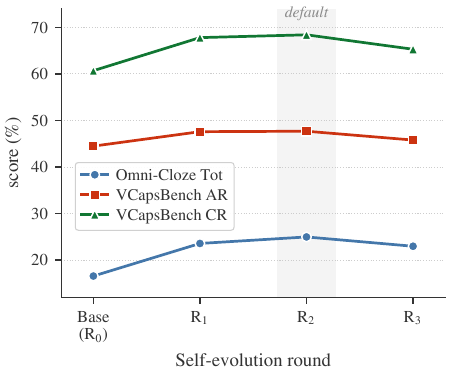}
\caption{Self-evolution trajectory across rounds. Round~$0$ is the base Qwen2.5-Omni-7B. \rev{We train two on-policy rounds as the default and additionally report a third round.} All three caption-fidelity metrics improve through Round~$2$ and then mildly regress at Round~$3$, consistent with a gradual saturation of the audit as the student improves.}
\label{fig:rounds_plateau}
\end{figure}

\subsection{Phase~A Audit Dimension Taxonomy ($D{=}\rev{29}$)}
\label{ssec:appendix-audit-taxonomy}

The dual-stream audit in Phase~A is organized around a fixed taxonomy of $D{=}\rev{29}$ audio-visual dimensions, designed to give the QA generator a stable, enumerable set of facets to check against. The taxonomy is drawn from the evaluation dimensions covered by VCapsBench~\citep{vcapsbench2025} on the visual and camera/aesthetic side, together with \rev{an audio and audio-visual taxonomy we define below}, but is intentionally coarser than either grid: a tight set of \rev{29} dimensions keeps both QA streams short enough to fit into a single audit pass while still spanning the perceptual aspects on which Video LLMs tend to drop content. We list the dimensions in the exact snake-case naming used by our QA-generator system prompt so that this appendix and the \rev{audit prompts} can be cross-referenced one-to-one. The \rev{29} dimensions are grouped into 5 macro categories.

\paragraph{Visual content (15).}
(1)~\texttt{subject\_existence}: which entities, people, or objects appear, their counts, and when they enter or leave; (2)~\texttt{subject\_attr}: identity, age, gender, profession, body features, material, shape, size; (3)~\texttt{subject\_relation}: comparative, family, hierarchical, or identity relations between subjects; (4)~\texttt{action\_interaction}: fine-grained actions and agent-to-object or agent-to-agent interactions; (5)~\texttt{state\_change}: how subjects or objects change state over time; (6)~\texttt{spatial\_layout}: 2D positions, left/right/above/below, direction, orientation; (7)~\texttt{scene\_setting}: place, environment, indoor/outdoor, background cues; (8)~\texttt{emotion\_atmosphere}: emotions of people, overall mood and atmosphere; (9)~\texttt{world\_knowledge}: culturally or historically specific items, famous people, brands, proper nouns; (10)~\texttt{lighting\_color}: lighting style and overall color palette/tone; (11)~\texttt{fine\_detail}: small peripheral details, textures, decorations, logos, patterns; (12)~\texttt{clothing\_accessories}: clothing items and accessories of each visible person; (13)~\texttt{color\_specific}: precise color of each named object (not the overall palette); (14)~\texttt{counting\_precise}: exact integer counts for distinct subject types; (15)~\texttt{scene\_segments}: how many scenes/shots the video contains and what each one depicts.

\paragraph{Camera and aesthetic \rev{(5)}.}
(16)~\texttt{camera\_movement}: camera motion type (pan, tilt, zoom, tracking, dolly, handheld, static, orbit) and how it changes; (17)~\texttt{shot\_type}: shot scale and framing (close-up, medium, wide, establishing, POV, aerial, over-the-shoulder); \rev{(18)~\texttt{blur\_dof}: depth of field (shallow bokeh vs.\ deep focus), focus pull, and motion blur;} (19)~\texttt{color\_grading}: post-processing color treatment (warm, cool, vintage, vivid, muted, high-contrast, desaturated, tinted); (20)~\texttt{visual\_style}: production or artistic style (cinematic, documentary, first-person, animation, timelapse, UGC vlog, AI-generated).

\paragraph{On-screen text (2).}
(21)~\texttt{native\_ocr}: real-world text inside the scene (signs, handwriting, labels, packaging); (22)~\texttt{overlay\_text}: superimposed subtitles, captions, or graphic overlays, including their content, position, and duration.

\paragraph{Audio (5).}
(23)~\texttt{speech\_content}: what is spoken, language, dialect, speaker identity, and transcribable phrases; (24)~\texttt{voice\_attribute}: voice timbre, pitch, emotion, delivery style (whisper, shout, sung, narration); (25)~\texttt{audio\_event}: non-speech sounds (music, instrument, animal, object, mechanical, SFX) and their sources; (26)~\texttt{sound\_scene}: acoustic environment inferred from audio (reverberant indoor, open outdoor, underwater, $\ldots$); (27)~\texttt{av\_sync}: audio-visual correspondence, on-screen vs.\ off-screen sound, narration timing.

\paragraph{Cross-modal (2).}
(28)~\texttt{anti\_hallucination}: probe whether described elements are actually present, and which commonly-assumed elements are NOT; (29)~\texttt{content\_type}: video category (educational, film/TV, social-media UGC, commercial, documentary, ad).

Each dimension comes with a one-line ``what to ask'' specification in the QA-generator system prompt, which the Caption QA Stream and the Video QA Stream both consume to generate $q_i$. For the Caption QA Stream, the generator is additionally allowed to return \texttt{SKIP\_NOT\_IN\_CAPTION} on dimensions that the initial caption $c_s$ does not address; the dropped dimensions go to $S_{\text{skip}}$ and become the seed of the Video QA Stream pivot (Section~\ref{sec:method:phaseA}).

\paragraph{Audit pipeline statistics.}
Table~\ref{tab:audit_stats} summarizes what Phase~A actually produces when the audit is run over the $2{,}000$-video training set in our default Round~1 frozen-teacher configuration. We measure two complementary layers. The \emph{caption-revision} numbers (top block) are computed directly from the audit JSONL that Phase~A writes: how often the audit fires and how much the caption is actually rewritten. The \emph{caption-vs-audit} numbers (bottom block) ask a different question: for every probe $(c_s, q_i, \mathrm{winner}_i)$ that survives Self-Voting and enters $\mathcal{E}_{\text{rqb}}$, is the student caption $c_s$ \emph{consistent with} the voted audit answer $\mathrm{winner}_i$? We answer this by sending each triple to a strong LLM judge (Gemini-2.5-Pro), which returns one of \{\textsc{consistent}, \textsc{inconsistent}, \textsc{absent}, \textsc{uncertain}\}; the four fields below summarize the resulting distribution.

\smallskip
\noindent\emph{Field definitions.} (i) \emph{Videos with $c_r \neq c_s$}: fraction of videos for which the audit actually triggers a rewrite. (ii) \emph{Mean character-level diff}: $|c_s - c_r|$ as a fraction of $|c_s|$, computed by Python's \texttt{difflib} ratio. (iii) \emph{Caption-present rate}: judge marked the caption as making a claim on that dimension (\textsc{consistent} or \textsc{inconsistent}). (iv) \emph{Caption-correct $|$ present}: among those, fraction marked \textsc{consistent}; this is the precision of what the caption already says. (v) \emph{Caption-commission rate}: fraction of \emph{all} probes marked \textsc{inconsistent}; the caption made a claim but the audit disagreed. (vi) \emph{Caption-absent rate}: fraction marked \textsc{absent}; the caption was silent on that dimension. By construction, present $+$ absent $+$ uncertain $\approx 1$.

\begin{figure}[t]
\centering
\includegraphics[width=\columnwidth]{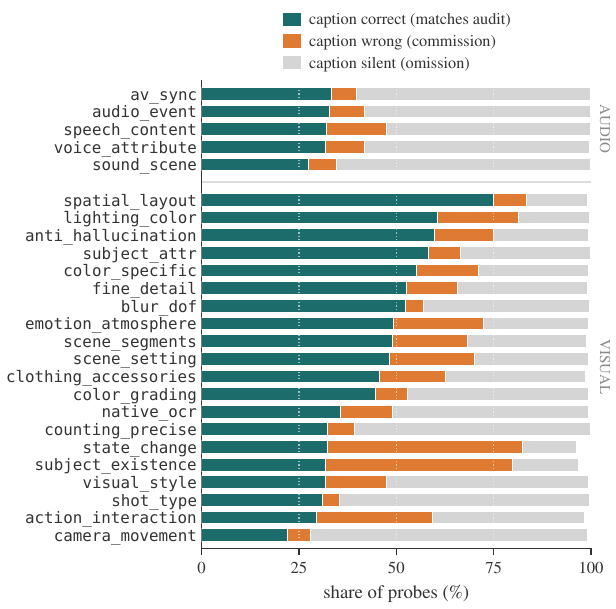}
\caption{Per-dimension caption-vs-audit decomposition from Table~\ref{tab:audit_stats} (Gemini-2.5-Pro judge). For each of the $25$ active dimensions, the stacked bar shows the share of probes where the initial caption $c_s$ is correct (teal, matches the voted audit answer), wrong (orange, commission), or silent (gray, omission). Rows are grouped by modality (audio block on top) and sorted within each group by correct-rate.}
\label{fig:audit_per_dim}
\end{figure}

\smallskip
\noindent\emph{What the numbers say.} Two patterns are worth flagging. (i)~The audit triggers an actual rewrite on a clear majority of videos, while keeping the median edit small, consistent with the design intent of a surgical revision step rather than a full rewrite. (ii)~Where the audit really earns its keep is precision and recall jointly: the judge marks a modest share of probes as \emph{inconsistent} with the voted audit, and a much larger share as \emph{absent}, with the absent share far higher on audio than on visual. This confirms the design motivation for the Video QA Stream: the caption leaks roughly a third of visual dimensions and well over half of audio ones, and a caption-only audit would miss those by construction. Figure~\ref{fig:audit_per_dim} shows the per-dimension decomposition. The audio block at the top is dominated by absent (gray) bars, confirming that omission, not commission, is the bottleneck on the audio side. Within the visual block, dimensions that require describing high-level facts the model is well-trained on (\texttt{spatial\_layout}, \texttt{lighting\_color}) are mostly correct, whereas dimensions that demand commitment to fine factual content (\texttt{state\_change}, \texttt{subject\_existence}, \texttt{action\_interaction}) accumulate the largest commission stripes, and camera-related dimensions (\texttt{camera\_movement}, \texttt{shot\_type}) are silent on a clear majority of clips. The figure makes the audit's design tractable: a single training pass extracts both the unwritten audio content and the over-confident state-change claims, two failure modes that the caption alone never surfaces.

\begin{table}[t]
\centering
\small
\resizebox{\columnwidth}{!}{%
\begin{tabular}{lc}
\toprule
\textbf{Quantity} & \textbf{Value} \\
\midrule
\multicolumn{2}{l}{\textit{Caption-level Revision}} \\
Videos with $c_r \neq c_s$                                      & $61.0\%$ \\
Mean character-level diff between $c_s$ and $c_r$               & $3.2\%$ \\
\midrule
\multicolumn{2}{l}{\textit{Caption-Audit Consistency}} \\
Caption-present rate, overall                                   & $57.2\%$ \\
\quad visual / audio                                            & $63.2\%$ / $41.2\%$ \\
Caption-correct $|$ present, overall                            & $71.2\%$ \\
\quad visual / audio                                            & $70.1\%$ / $75.7\%$ \\
Caption-commission rate, overall                                & $16.5\%$ \\
\quad visual / audio                                            & $18.9\%$ / $10.0\%$ \\
Caption-absent rate, overall                                    & $41.9\%$ \\
\quad visual / audio                                            & $35.7\%$ / $58.6\%$ \\
\bottomrule
\end{tabular}%
}
\caption{Phase~A audit statistics on the $2{,}000$-video training set, Round~1 (frozen-teacher default). The top block is computed directly from the audit JSONL; the bottom block reports an LLM-as-judge (Gemini-2.5-Pro) assessment over the surviving probes of whether $c_s$ is consistent with the audit's voted answer.}
\label{tab:audit_stats}
\end{table}

\subsection{Supplementary Caption Evaluation with VCapsBench-Omni}
\label{ssec:appendix-vcapsbench-omni}

Omni-Cloze and UGC-VideoCap provide complementary measures of caption fidelity. For our experiments in Section~\ref{sec:exp}, we also needed question-level checks of specific sounds and how they relate to visible sources or actions. We therefore extend a subset of VCapsBench~\citep{vcapsbench2025} with audio and joint audio-visual questions, yielding the supplementary evaluation set \emph{VCapsBench-Omni}. We adapt the original VCapsBench construction pipeline (Figure~4 of \citealp{vcapsbench2025}) in three ways.
\begin{enumerate}
    \item the visual yes/no QA set is reused \emph{verbatim} from VCapsBench (no re-generation, no re-curation), so that the visual track of VCapsBench-Omni is comparable, video-by-video and QA-by-QA, with the original benchmark;
    \item we add two new tracks on top of the visual one, namely an \emph{audio} track (4~macro / 13~sub categories) and an \emph{audio-visual joint} track (1~macro / 5~sub categories), and use the same two-pipeline-plus-correction protocol to construct them;
    \item all VLM and LLM calls in the new tracks are issued to a single annotator model, \texttt{Gemini~2.5~Pro}, which acts both as the multimodal annotator (Pipelines~1 and~2) and as the textual curator (Data Correction Pipeline).
\end{enumerate}
The remainder of this appendix describes the taxonomy, the three pipelines, and the final merging step. All prompts and QA strings are in English.

\subsubsection{Taxonomies}
\label{ssec:appendix-taxonomy}

The visual taxonomy is identical to VCapsBench: 4 macro categories and 21 sub categories covering action, count, text, color, position, relative position, entity, entity shape, entity size, background, relation, lighting, style, shot type, atmosphere, camera movement, color palette, color grading, blur, emotion, and proper noun. We do not modify it.

For audio, we \rev{use the following} 4 macro categories with 13 sub categories:
\begin{itemize}
    \item \textbf{Human\_Voice\_Speech}: Speaking\_State, Non\_Verbal\_Vocalization;
    \item \textbf{Physical\_SFX}: Action\_SFX, Rigid\_Body\_Collision, Fluid\_Sound, Mechanical\_Sound, Percussion\_SFX;
    \item \textbf{Ambience}: Natural\_Ambience, Social\_Ambience, Silence;
    \item \textbf{Background\_Music}: Emotional\_Tone, Music\_Genre, Rhythm\_Tempo.
\end{itemize}

For the audio-visual joint track, we restrict ourselves to a single macro category (\textbf{AV\_Joint}) with 5 sub categories that all require evidence from \emph{both} modalities to be answered: \emph{Sound\_Source\_Grounding}, \emph{Lip\_Sync}, \emph{Action\_SFX\_Alignment}, \emph{AV\_Emotion\_Consistency}, and \emph{Off\_Screen\_Sound}. We deliberately exclude acoustic-temporal questions (precise sound onset / offset, event count) because current models, including the strongest closed-source ones, are unreliable on those.

\subsubsection{Pipeline 1: Caption-then-QA}
\label{ssec:appendix-p1}

Pipeline~1 mirrors the upper branch of Figure~4 of \citet{vcapsbench2025} and is split into two stages.

\paragraph{Stage 1.1: detailed audio + AV-binding caption.}
For every video, Gemini~2.5~Pro is asked to watch the clip and emit one detailed English caption that thoroughly covers the audio content and the audio-visual binding. Visual-only details are deliberately kept brief, since the visual side is already covered by VCapsBench's native QA. The model is instructed to (i)~enumerate every salient sound (human voice, physical SFX, ambience, music) and its perceptual attributes (pace, tone, emotion, genre, tempo); (ii)~bind salient sounds to their visible emitters whenever possible (source localization, lip sync, action-to-SFX alignment, AV emotion consistency); and (iii)~mark off-screen sounds explicitly. The caption is constrained to 80 to 200 plain-prose English words, with no hedging language (``maybe'', ``probably''), and is returned as a STRICT-JSON object with a single \texttt{caption} field.

\paragraph{Stage 1.2: LLM-mined yes/no QAs from the caption.}
The captions from Stage~1.1 are then fed, video-by-video, to Gemini~2.5~Pro \emph{in text-only mode}. The model is asked to mine two kinds of yes/no QA-pairs that must be derivable from the caption alone: \texttt{audio\_qa} (answerable from the audio content described in the caption) and \texttt{av\_joint\_qa} (require both an audio and a visual cue described in the caption). The prompt enforces several hard constraints:
\begin{itemize}
    \item Each \texttt{question} is a binary yes/no question and \texttt{answer} is exactly ``yes'' or ``no''; the LLM is forbidden from inventing facts not present in the caption.
    \item Each audible event is decomposed into multiple atomic QAs covering existence / type / attribute / source binding.
    \item For every salient verifiable absence in the caption (e.g.\ ``no music''), one or two negative (\texttt{answer:no}) audio QAs are produced.
    \item Every \texttt{av\_joint\_qa} question must \emph{simultaneously} reference a visual cue and an audio cue; the only valid AV-joint types are the five listed in Section~\ref{ssec:appendix-taxonomy}.
    \item Per-video budgets are 5 to 10 audio QAs and 4 to 8 AV-joint QAs; near-silent clips emit zero QAs.
    \item Categories are picked strictly from the audio and AV-joint taxonomies, using the exact English tags.
\end{itemize}
The LLM emits a STRICT-JSON object with two fields, \texttt{audio\_qa} and \texttt{av\_joint\_qa}, each entry carrying \texttt{(concept, question, answer, category\_1, category\_2)}.

\subsubsection{Pipeline 2: Direct QA from the Video}
\label{ssec:appendix-p2}

Pipeline~2 mirrors the lower branch of Figure~4 of \citet{vcapsbench2025}: instead of going through an intermediate caption, Gemini~2.5~Pro watches the video directly and emits yes/no QAs grounded in what it can see and hear. The taxonomy and per-category budgets are identical to those in Pipeline~1; the same atomic-decomposition and salient-absence rules apply, and the same hard constraint (every AV-joint question must reference a visual \emph{and} an audio cue) is enforced. The resulting QAs play the same role as the corresponding lower branch in the original VCapsBench Figure~4, providing an independent source that is later cross-checked against Pipeline~1.

\subsubsection{Data Correction Pipeline}
\label{ssec:appendix-correct}

The QA-pool produced by merging Pipelines~1 and~2 is cleaned in two stages, exactly mirroring the data-correction pipeline in Figure~4 of \citet{vcapsbench2025}.

\paragraph{Stage A: rule-based pre-filter.}
For every QA in the merged pool we record a \texttt{sources} tag indicating which of $\{p_1, p_2\}$ produced it, and a \texttt{per\_source\_answers} list with the yes/no answer each pipeline gave. The rule layer is deterministic and cheap:
\begin{enumerate}
    \item \emph{Conflict drop}: any QA whose \texttt{per\_source\_answers} contains both ``yes'' and ``no'' is dropped (always on);
    \item \emph{No-concept drop}: any QA with an empty \texttt{concept} field is dropped (on by default; suggests low-quality generation);
    \item \emph{Singleton drop} (optional): any QA produced by only one of the two pipelines is dropped, mirroring the ``appears only once'' filter described in the original VCapsBench paper.
\end{enumerate}

\paragraph{Stage B: LLM curator.}
For every \texttt{(video, category\_2)} bucket, the rule-filtered QAs are passed to Gemini~2.5~Pro acting as a senior data curator. The curator is asked to:
\begin{enumerate}
    \item \emph{Merge near-duplicates.} If two questions ask the same underlying audio fact in different wording, keep exactly one (preferring the clearer / more specific phrasing).
    \item \emph{Drop low-quality items.} Drop QAs whose questions are vague, ambiguous, unanswerable from audio/AV evidence, or near-tautological.
    \item \emph{Preserve ground-truth answers as given.} Yes/no flips are explicitly forbidden.
    \item \emph{Respect the category.} QAs cannot be moved to a different category.
\end{enumerate}
Bucket sizes are small by construction (one video $\times$ one sub category), so each curator call is cheap. On curator failure we fall back to the rule-filtered output for that bucket.

\subsubsection{Final Merge and Format}
\label{ssec:appendix-merge}

Three sources are merged to produce the final dataset: (i)~the video list from a balanced 1k subset of VCapsBench, (ii)~the visual yes/no QAs taken \emph{verbatim} from \texttt{VCapsBench\_100KQA.jsonl}, and (iii)~the audio and AV-joint QAs produced by the data-correction pipeline. Each output line carries the visual QAs in the original VCapsBench triple-string format, the structured audio QAs, the structured AV-joint QAs, and ``flat'' mirrors that re-encode the audio and AV-joint QAs into the same triple-string format so that any VCapsBench-style evaluator can also consume them directly. The audio caption from Pipeline~1 is retained as a reference field but is not part of the evaluation set.

\subsubsection{Per-model Evaluation Protocol}
\label{ssec:appendix-eval}

Per-model evaluation also follows the standard VCapsBench protocol. The model under test first emits one English omni caption per video. An LLM judge (also Gemini~2.5~Pro in our setup) then verifies each yes/no QA \emph{solely from the caption text}; the eval scripts never re-watch the video. Each verdict is one of $\{$\texttt{Yes}, \texttt{No}, \texttt{Unanswerable}$\}$, mapped to the three standard rates: accuracy rate (AR$\uparrow$, fraction of QAs answered correctly), inconsistency rate (IR$\downarrow$, fraction answered incorrectly), and coverage rate (CR$\uparrow{=}1{-}\text{uncertain rate}$, fraction the caption contains enough information to answer at all). The judge prompts are slightly track-specific: the visual judge is the standard ``answer yes / no / unanswerable based only on caption'' template; the audio judge prioritises the core audio event referenced by the question and ignores non-audio frills; the AV-joint judge is stricter and requires the caption to explicitly mention the visual fact, the audio fact, and the binding between them, otherwise the verdict is \texttt{Unanswerable}. The three rates are reported per track (visual / audio / AV-joint) and as a QA-count-weighted overall score.

\subsection{Prompt Templates for Phase~A and Phase~B}
\label{ssec:appendix-prompts}

This subsection makes the two prompt templates used throughout Section~\ref{sec:method} explicit. The same two strings are reused across phases: the captioning prompt $\mathrm{prompt}_c$ conditions the student in Phase~A draft generation (Eq.~\ref{eq:student_caption}) and the student path in Phase~B (Eq.~\ref{eq:student_logits}); the revision prompt $\mathrm{prompt}_r(c_s, \mathcal{E}_{\text{rqb}})$ conditions the teacher in Phase~A revision (Eq.~\ref{eq:teacher_caption}) and the teacher path in Phase~B (Eq.~\ref{eq:teacher_logits}). Both are fixed strings; only $c_s$ and $\mathcal{E}_{\text{rqb}}$ are spliced in at runtime. We show the bare-string content below; the backbone's standard chat template wraps each block as a single user message.

\paragraph{Captioning prompt $\mathrm{prompt}_c$.}
Deliberately minimal so that the student is forced to scan the audio-visual input without any text-side hint about which dimensions to cover.
\begin{quote}\small\ttfamily
You are given a short video with an audio track. Write one detailed English caption that describes both what you see and what you hear in the video, as a single paragraph of 80-200 words. Cover salient subjects, actions, on-screen text, camera work, background music, speech content, ambient sounds, and any audio-visual binding (who is speaking, what makes a given sound). State only what the video supports. Do not use hedging language such as ``maybe'' or ``possibly''.
\end{quote}

\paragraph{Revision prompt $\mathrm{prompt}_r(c_s, \mathcal{E}_{\text{rqb}})$.}
Splices the student draft $c_s$ and the verified bank $\mathcal{E}_{\text{rqb}}$ (rendered as a numbered list of $(q_i, \hat a_i)$ pairs) into a fixed shell of revision principles. Angle-bracketed tokens mark per-example splice points; \texttt{<c\_s>} is replaced by the verbatim text of the initial caption and each \texttt{<q\_i>}\,/\,\texttt{<a\_i>} by the $i$-th surviving QA pair from $\mathcal{E}_{\text{rqb}}$.
\begin{quote}\small\ttfamily
You are revising a video caption.\\[2pt]
[INITIAL CAPTION]\\
<c\_s>\\[2pt]
[VERIFICATION Q\&A]\\
The following questions have been verified against the audio-visual content of the video. Each answer is the verified ground truth and takes FINAL authority over the initial caption.\\
~~1. Q: <q\_1> ~$\to$~ A: <a\_1>\\
~~2. Q: <q\_2> ~$\to$~ A: <a\_2>\\
~~~~~$\cdots$\\
~~N. Q: <q\_N> ~$\to$~ A: <a\_N>\\[2pt]
[INSTRUCTIONS]\\
- Rewrite the initial caption into a single revised caption that is consistent with every verified answer above.\\
- Keep the spans that the verification confirms; rewrite only the spans that the verification contradicts or under-specifies.\\
- For any audio question whose answer is ``none'', remove the corresponding audio claim from the caption (do not assert that sound).\\
- Do not introduce facts that are neither in the initial caption nor in the verification list.\\
- Do not use hedging language such as ``maybe'' or ``possibly''.\\
- Return one English caption of 80-200 words as a single paragraph.
\end{quote}

\paragraph{How Phase~B consumes these prompts.}
At Phase~B training time, both prompts are rendered into the textual context that prefixes the same teacher-forced caption $c_r$. For each cached tuple $(c_s, \mathcal{E}_{\text{rqb}}, c_r)$, the student path encodes \rev{the low-budget view $V_{\text{lo}}$} and prefixes $\mathrm{prompt}_c$; the teacher path encodes \rev{the high-budget view $(V_{\text{hi}}, A_{\text{hi}})$} and prefixes $\mathrm{prompt}_r(c_s, \mathcal{E}_{\text{rqb}})$ under \texttt{torch.no\_grad}. Both paths then read off logits over the same $c_r$ token positions, and Eq.~\ref{eq:opd_loss} averages the per-token KL between them over $|c_r|$. \rev{The two paths differ in both the textual prefix and the input budget, so the loss carries the text-side lift induced by $(c_s, \mathcal{E}_{\text{rqb}})$ together with the multimodal lift}; the splice points in the revision prompt are re-evaluated only when Phase~A is refreshed (every $R$ optimization steps), and are otherwise read straight from the cache.

\subsection{Phase A Walkthrough: A Real Example}
\label{ssec:appendix-phaseA-example}

To make the dual-stream audit and the revision step concrete, we walk through one real training example produced by Phase~A, copied verbatim from the cached tuple $(c_s, \mathcal{E}_{\text{rqb}}, c_r)$. The example uses an in-the-wild silent video showing a finger pointing at two leaves on the soil. We display three items in turn: the student draft $c_s$, the QA pairs that survived Self-Voting (i.e.\ the contents of $\mathcal{E}_{\text{rqb}}$), and the revised caption $c_r$ that the teacher writes conditioned on $c_s$ and $\mathcal{E}_{\text{rqb}}$.

\paragraph{Student caption $c_s$ (low-budget view).}
\begin{quote}\small\itshape
The video opens with a close-up shot of two green leaves resting on a patch of soil. The soil appears to contain some dried plant matter and small pieces of wood. A few green plants are also visible in the background. A finger, likely belonging to the person holding the camera, points at the left leaf first and then the right leaf. The camera then zooms in for a closer look at the leaves, showcasing their texture and the veins running through them. The finger continues to point at both leaves, emphasizing specific areas. The camera remains focused on the leaves throughout, with the background remaining constant. There are no other people or objects in the frame. The video seems to be a detailed examination of the leaves, possibly for educational or observational purposes. The lighting is natural, suggesting an outdoor setting. There is no audio present in the video.
\end{quote}

\paragraph{Revision QA Bank $\mathcal{E}_{\text{rqb}}$.}
We list each QA pair in $\mathcal{E}_{\text{rqb}}$ in the form ``$q$ $\to$ $\hat a$ ($\rho$)''. Each $\hat a$ is the majority answer over $K{=}5$ teacher samples on the high-budget view, and $\rho$ is the corresponding agreement rate; only pairs with $\rho{\ge}\tau$ and $\hat a\!\neq\!\text{``uncertain''}$ are retained.
\begin{itemize}\setlength{\itemsep}{1pt}\small
\item \textbf{Q1} (visual): \emph{What texture do the leaves have?} \hfill $\to$ \emph{smooth} ($\rho{=}0.67$).
\item \textbf{Q2} (visual): \emph{What type of environment is the video shot in?} \hfill $\to$ \emph{outdoor} ($\rho{=}1.00$).
\item \textbf{Q3} (visual): \emph{What is the color of the soil?} \hfill $\to$ \emph{brown} ($\rho{=}0.67$).
\item \textbf{Q4} (\texttt{blur\_dof}): \emph{What is the depth of field in the video?} \hfill $\to$ \emph{shallow} ($\rho{=}0.67$).
\item \textbf{Q5} (audio): \emph{Is there any background music audible in the video?} \hfill $\to$ \emph{none} ($\rho{=}0.67$).
\item \textbf{Q6} (audio): \emph{Is there any spoken dialogue in the video?} \hfill $\to$ \emph{none} ($\rho{=}1.00$).
\item \textbf{Q7} (audio): \emph{Are there any non-music sound effects in the video?} \hfill $\to$ \emph{none} ($\rho{=}1.00$).
\item \textbf{Q8} (audio): \emph{Is there any narration in the video?} \hfill $\to$ \emph{none} ($\rho{=}1.00$).
\end{itemize}
Two patterns are worth noting. First, Q1 to Q4 are visual-content checks that fill in fine-grained attributes the student left implicit (\emph{smooth} texture, \emph{brown} soil, \emph{shallow} depth of field), produced by the Caption QA Stream from $c_s$. Second, Q5 to Q8 are audio negatives with $\hat a{=}\text{none}$; these are the kind of dimensions $c_s$ tends to omit and that the Video QA Stream is designed to recover. They survive Self-Voting at $\rho{\ge}0.67$ because the teacher consistently reports silence across all $K$ samples on the high-budget view.

\paragraph{Teacher prompt for revision.}
The teacher receives the student draft $c_s$ followed by $\mathcal{E}_{\text{rqb}}$ as a numbered ``Verification Q\&A'' list, plus a fixed set of revision principles (greedy decoding, single pass, ground-truth answers have FINAL authority over $c_s$, hedging words removed, audio claims with $\hat a_i{=}\texttt{none}$ must be removed). The full template is the revision prompt $\mathrm{prompt}_r(c_s, \mathcal{E}_{\text{rqb}})$ given in Appendix~\ref{ssec:appendix-prompts}.

\paragraph{Revised caption $c_r$ (teacher, high-budget view).}
\begin{quote}\small\itshape
The video begins with a close-up shot of two green leaves resting on a patch of soil. The soil appears to contain some dried plant matter and small pieces of wood. A few green plants are also visible in the background. A finger, likely belonging to the person holding the camera, points at the left leaf first and then the right leaf. The camera then zooms in for a closer look at the leaves, showcasing their \textbf{smooth texture} and the veins running through them. The finger continues to point at both leaves, emphasizing specific areas. The camera remains focused on the leaves throughout, with the background remaining constant. There are no other people or objects in the frame. The video seems to be a detailed examination of the leaves, possibly for educational or observational purposes. The lighting is natural, suggesting an outdoor setting. There is no audio present in the video.
\end{quote}
The teacher keeps the parts of $c_s$ that the audit confirmed (e.g.\ ``natural lighting'', ``no audio''), and rewrites only the spans contradicted or under-specified by $\mathcal{E}_{\text{rqb}}$: in this example, the only edit is promoting ``their texture'' to ``their \emph{smooth} texture'', i.e.\ folding the answer of Q1 into $c_r$. Q2 to Q4 happen to be already implicitly consistent with $c_s$, while the four audio negatives Q5 to Q8 are absorbed into the unchanged closing sentence ``There is no audio present in the video.''. This is the typical behavior we observe across the corpus: most QA pairs serve as silent verifications of $c_s$, and a small number trigger localized edits that thread the audit's evidence into the revised text.

\paragraph{What Phase~B sees.}
At training time, this single example contributes one cached tuple $(c_s, \mathcal{E}_{\text{rqb}}, c_r)$. The teacher path is teacher-forced to $c_r$ with the textual context $(c_s, \mathcal{E}_{\text{rqb}})$ rendered as the numbered Q\&A list above; the student path is teacher-forced to the same $c_r$ but conditioned only on $\mathrm{prompt}_c$. The KL in Eq.~\ref{eq:opd_loss} is then the per-token gap between these two distributions over the caption tokens of $c_r$, so the supervision is densest precisely on the spans where the audit injected new content (e.g.\ ``smooth'').

Algorithm~\ref{alg:opd} summarizes the full procedure. Phase~A is executed once at the start and then every $R$ optimization steps; between refreshes, Phase~B replays the cached $(c_s, \mathcal{E}_{\text{rqb}}, c_r)$ with gradients enabled.

\begin{algorithm}[t]
\caption{SODA: Self-Supervised On-Policy Distillation via Dual-Stream Auditing.}
\label{alg:opd}
\begin{algorithmic}[1]
\Require model $M_\theta$; training set $\mathcal{D}$; refresh period $R$; vote size $K$, thresholds $\tau, \tau_{\text{piv}}$; pivot cap $D_{\max}$
\State \textbf{Phase A.} For each $(V,A) \in \mathcal{D}$:
\State \quad sample $c_s \sim M_\theta(V_{\text{lo}}, \mathrm{prompt}_c)$ \Comment{student draft}
\State \quad \emph{Caption QA Stream:} $\{(q_i, d_i, m_i)\} \gets \mathrm{QGen}(c_s)$;
\State \quad \quad $S_{\text{skip}} \gets \{d_i : q_i {=} \texttt{SKIP\_NOT\_IN\_CAPTION}\}$
\State \quad \quad for each remaining $q_i$: sample $K$ answers from $M_\theta(V_{\text{hi}}, A_{\text{hi}}, q_i)$, take majority $(\hat a_i, \rho_i)$
\State \quad \quad $\mathcal{E}_{\text{rqb}}^{\text{cap}} \gets \{(q_i, \hat a_i) : \rho_i \!\ge\! \tau,\, \hat a_i {\neq} \text{uncertain},\, d_i {\notin} S_{\text{skip}}\}$
\State \quad \emph{Video QA Stream (video-grounded pivot):} $U \gets S_{\text{skip}} \cup \{d_i {:} \rho_i {<} \tau_{\text{piv}}\text{ or dropped}\}$, capped at $D_{\max}$
\State \quad \quad $\{q_j\} \gets \mathrm{QGen}(V_{\text{hi}}, A_{\text{hi}}, U; \,\text{avoid}{=}\{q_i\})$
\State \quad \quad apply the same $K$-sample vote to each $q_j$ to obtain $\mathcal{E}_{\text{rqb}}^{\text{vid}}$
\State \quad $\mathcal{E}_{\text{rqb}} \gets \mathcal{E}_{\text{rqb}}^{\text{cap}} \oplus \mathcal{E}_{\text{rqb}}^{\text{vid}}$
\State \quad $c_r \gets \arg\max\, M_\theta(V_{\text{hi}}, A_{\text{hi}}, c_s, \mathcal{E}_{\text{rqb}})$
\State \quad cache $(c_s, \mathcal{E}_{\text{rqb}}, c_r)$
\State \textbf{Phase B.} For each step $t=1,2,\dots$:
\State \quad sample a cached tuple $(c_s, \mathcal{E}_{\text{rqb}}, c_r)$
\State \quad $\boldsymbol{\ell}_s \gets M_\theta(V, \mathrm{prompt}_c;\ \mathrm{forced}{=}c_r)$
\State \quad $\boldsymbol{\ell}_t \gets M_\theta(V, c_s, \mathcal{E}_{\text{rqb}};\ \mathrm{forced}{=}c_r)$ \Comment{no gradient; same audio-visual encoding as student}
\State \quad update $\theta$ by minimizing Eq.~\ref{eq:opd_loss}
\State \quad if $t \bmod R = 0$, re-run Phase~A
\end{algorithmic}
\end{algorithm}

\subsection{Qualitative Comparison: Base vs.\ SODA-trained Student}
\label{ssec:appendix-qualitative}

To complement the aggregate caption-fidelity numbers in Section~\ref{sec:exp:caption}, we walk through one Omni-Cloze sample for which the SODA-trained student (frozen-teacher Round~2) writes a noticeably more faithful caption than the base Qwen2.5-Omni-7B. Concrete details that appear in the SODA caption \emph{and} also surface in the ground-truth Omni-Cloze passage but are missing from the base caption are highlighted in \textcolor{green!50!black}{\textbf{green}}. We deliberately keep this comparison qualitative; the corresponding aggregate numbers are reported in Table~\ref{tab:caption}.

\paragraph{Sample: Comedic monologue in a living room.} A woman delivers a sketch monologue while a dog yawns on the couch behind her; bold captions appear in sync with her speech.

\paragraph{Base caption.}
\begin{quote}\small\itshape
The video opens with a woman standing in a cozy living room. \dots{} A coffee table sits in front of the couch, holding a few books and a small potted plant. To the right of the couch, there is a bookshelf filled with books and some decorative items. \dots{} As she walks towards the camera, she starts talking. The on-screen text appears, reading ``OH, SORRY. DID YOU EXPECT ME TO FLINCH? TO HAVE A REACTION?'' \dots{} \textbf{a large black and brown dog is seen lying on the couch}. \dots{} The audio quality is clear, allowing the viewer to easily understand the woman's speech and the on-screen text.
\end{quote}

\paragraph{SODA caption.}
\begin{quote}\small\itshape
The video opens with a woman standing in a living room. \dots{} A small table in front of the couch holds a few items, including \textcolor{green!50!black}{a cup} and a book. \textcolor{green!50!black}{A bookshelf filled with books is visible in the background, along with a framed picture on the wall.} \textcolor{green!50!black}{A large dog is lying on the couch, looking relaxed.} \dots{} \textcolor{green!50!black}{The on-screen text appears in yellow and white,} saying ``OH. SORRY. DID YOU EXPECT ME TO FLINCH. TO HAVE A REACTION? \dots\,''.
\end{quote}

The SODA caption recovers a cup on the coffee table, a framed picture on the wall, and crucially the colour scheme of the on-screen captions (yellow and white), all of which are anchored as cloze blanks in the ground-truth passage but absent from the base caption. The SODA caption is also more cautious about the dog: it reports a large dog \emph{looking relaxed} rather than committing to the unverified \emph{black and brown} colour that the base caption invents.

\subsection{Extended Discussion of Related Work}
\label{ssec:appendix-related-works}

This appendix gives the longer version of the three threads briefly summarized in Section~\ref{sec:related}.

\subsubsection{On-Policy Distillation}
\label{ssec:appendix-related-opd}

Knowledge distillation~\citep{hinton2015distilling} trains a student to imitate a teacher's output distribution, and has long been a backbone of compressing language models. Sequence-level distillation~\citep{kim2016seqkd} extends this idea to autoregressive generation by aligning whole-sequence likelihoods, while $f$-divergence variants~\citep{f-distill2024,minillm2024} replace the forward-KL with reverse-KL or Jensen-Shannon to reduce mode-averaging in language generation. A central problem in this line is the \emph{train-inference mismatch}: distillation traditionally fits the student's distribution at the teacher's outputs, while at deployment the student decodes from its own. On-policy distillation~\citep{gkd2024,distillm2024} addresses this by sampling from the student during training and computing the teacher's likelihood on those samples, so that the supervision signal lives on the trajectories the student actually visits. G-OPD~\citep{gopd2026} ports this paradigm to multimodal LLMs by using a stronger external text-only model as the teacher and aligning student outputs along the text channel.

This entire family relies on a teacher that is \emph{stronger} than the student in the standard sense: more parameters, data, or compute. SODA explores a different point: the teacher and student share weights, and the teacher's edge comes from a richer input rather than a larger model, so the supervision reflects an information gap rather than a capacity gap and removes the dependence on an external stronger model.

\subsubsection{Self-Distillation and Self-Training}
\label{ssec:appendix-related-selfdistill}

A separate line of work uses a model to teach itself. Born-again networks~\citep{bornagain2018} retrain a student of the same architecture on the previous generation's outputs and observe consistent gains. Be-Your-Own-Teacher~\citep{byot2019} distills deep layers into shallow ones inside a single network. In semi-supervised learning, mean teachers~\citep{meanteacher2017} and noisy student~\citep{noisystudent2020} construct the teacher from a stable copy of the student plus stronger augmentations on the unlabeled side. For language reasoning, STaR~\citep{stardb2023} and ReST~\citep{rest2024} bootstrap a model by sampling rationales or solutions from itself, filtering with a correctness oracle, and fine-tuning on the survivors; self-consistency~\citep{wang2023selfconsistency} aggregates multiple samples into a single, more reliable answer. Recent video-side work such as Video-STaR~\citep{videostar2024} extends this self-training idea to instruction tuning.

These methods typically separate teacher and student along a \emph{noise} or \emph{stochasticity} axis (augmentations, EMA, majority voting) and supervise on the final answer, which provides only sparse learning signal. SODA instead separates them along an \emph{information} axis, where the teacher sees a higher-fidelity audio-visual view together with the student's draft and a model-produced Revision QA Bank, and supervises on caption tokens, turning text-side and multimodal-side input-budget lifts into a dense token-level signal for an omni-modal VideoLLM.

\subsubsection{Reducing Hallucinations in Video and Multimodal LLMs}
\label{ssec:appendix-related-hallu}

Hallucination is a recognized failure mode of multimodal LLMs and has been studied along several axes. On the evaluation side, POPE~\citep{popebench2023} and HallusionBench~\citep{hallusionbench2024} target object-level and entanglement-level hallucinations in image LLMs, and VideoHallucer~\citep{videohallucer2024}, VidHalluc~\citep{vidhalluc2025}, EventHallusion~\citep{eventhallusion2025}, and AVHBench~\citep{avhbench2025} extend the diagnosis to videos and to audio-visual interactions. On the mitigation side, three families dominate. \emph{Decoding-time interventions} such as VCD~\citep{vcd2024} and OPERA~\citep{opera2024} contrast the model's distribution under perturbed inputs to suppress over-confident hallucinations. \emph{Preference-based fine-tuning} methods including RLHF-V~\citep{rlhfv2024}, POVID~\citep{povid2024}, HA-DPO~\citep{halva2024}, and LURE~\citep{lure2024} collect or synthesize hallucination-free vs.\ hallucinated pairs and apply DPO~\citep{dpo2023} or RLHF on top of the answer tokens. \emph{Reinforcement-style methods} such as SDRL~\citep{sdrl2026} and VideoHallu~\citep{videohallu2025} optimize answer accuracy or hallucination metrics with GRPO~\citep{grpo2024,ppo2017}. A complementary thread uses an external process or reward model, e.g.\ VisualPRM~\citep{visualprm2026}, to score intermediate reasoning steps.

These methods place supervision on answer tokens (or contrastive perturbations of them), often require curated preference data, reward models, or an RL loop on top of SFT, and provide a sparse signal, namely a binary label or a scalar reward per trajectory. SODA differs on all three axes: a dense token-level KL on the revised caption, a same-weight teacher with no preference data or reward model, and audio and video input-budget lifts treated symmetrically inside the teacher's input; we therefore view our gains on hallucination benchmarks as complementary to this literature rather than competing with it.

\end{document}